\documentclass[letterpaper,journal]{IEEEtran}

\usepackage[utf8]{inputenc}
\usepackage[T1]{fontenc}
\usepackage{cite}
\usepackage{url}
\usepackage{booktabs}
\usepackage{amsfonts}
\usepackage{nicefrac}
\usepackage{microtype}
\usepackage{xcolor}
\usepackage{graphicx}
\usepackage{array}
\usepackage{textcomp}
\usepackage{stfloats}
\usepackage[caption=false,font=footnotesize]{subfig}
\usepackage{bm}
\usepackage{multirow}
\usepackage{longtable}
\usepackage{makecell}

\usepackage{amsmath}
\usepackage{amssymb}
\usepackage{mathtools}
\usepackage{amsthm}

\theoremstyle{plain}
\newtheorem{theorem}{Theorem}[section]
\newtheorem{proposition}[theorem]{Proposition}

\theoremstyle{definition}

\theoremstyle{remark}

\usepackage{titletoc}
\usepackage[hidelinks]{hyperref}
\usepackage[capitalize,noabbrev]{cleveref}

\usepackage[disable,textsize=tiny]{todonotes}

\begin{document}

\title{Exploring the Potential of Multi-objective Optimization \\ for Single-task Model Learning}

\author{Rasa Khosrowshahli, Brendan Park, Aws Al Jumaily, Tsz Yi Kwok, Qi Chen Gao, Maysara Al Jumaily, Beatrice Ombuki-Berman, Shahryar Rahnamayan, and Yifeng Li%
\thanks{Rasa Khosrowshahli, Brendan Park, Aws Al Jumaily, Tsz Yi Kwok, Qi Chen Gao, Maysara Al Jumaily, Beatrice Ombuki-Berman, and Yifeng Li are with the Department of Computer Science, Faculty of Mathematics and Science, Brock University, St. Catharines, ON L2S 3A1, Canada.}%
\thanks{Shahryar Rahnamayan is with the Department of Engineering, Faculty of Mathematics and Science, Brock University, St. Catharines, ON L2S 3A1, Canada.}%
\thanks{Corresponding authors are Rasa Khosrowshahli (rkhosrowshahli@brocku.ca) and Yifeng Li (yli2@brocku.ca).}}

\markboth{Preprint}%
{Khosrowshahli et al.: Exploring the Potential of Multi-Objective Optimization for Single-Task Model Learning}

\maketitle

\begin{abstract}
Single-task neural networks are often trained on a weighted sum of several criteria. Generative models balance reconstruction against latent regularization, self-supervised segmentation combines spatial and biological constraints, and reinforcement learning rewards aggregate several properties of one output. Even cross-entropy decomposes into class-conditional risks. The weighted sum of objectives reduces the objective-wise gradients to a single vector, so an update that minimizes the total objectives can maximize at least one of objectives. In this paper, we study when established multi-objective optimization (MOO) methods that aggregate objective-wise gradients improve on fixed linear scalarization (LS) in single-task learning. We express four applications in a common Jacobian notation and state a first-order condition under which an LS step increases an individual loss. We then compare LS with MGDA and its normalized variants, AMTL, UPGrad, and smooth Tchebycheff scalarization on long-tailed classification with up to 100 class-wise objectives, self-supervised cell segmentation with six losses, variational and vector-quantized autoencoders with two to five objectives, and molecular lead optimization with three rewards. Objective-wise optimization improved selected outcomes such as normalized MGDA reduced VAE reconstruction FID on CelebA from 79.0 to 46.6, and multi-objective reinforcement learning raised the share of molecules meeting all three property criteria from 23.5\% to 54.1\%. The gains depended strongly on the decomposition, normalization, and aggregation method. Class-balanced decomposition with LS accounted for most classification gains, and UPGrad improved a five-objective VQ-VAE but degraded the three-objective base model. Measured runtime overhead ranged from 2\% to 64\% over the corresponding scalarized baselines. We derive practical guidance on when this added cost is justified.
\end{abstract}


\section{Introduction}

\IEEEPARstart{S}{ingle-task} learning does not necessarily involve a single training criterion. A generative model may balance reconstruction and latent regularization, a segmentation model may combine spatial and biological constraints, and a reinforcement learning policy may optimize several properties of the same output. Even a classification criterion can be decomposed into class-conditional risks. These objectives are combined through a weighted sum before a gradient-based optimizer updates the model parameters. This approach is simple and often effective, but the resulting update may not improve every individual objective.

Multi-Objective Optimization (MOO) provides tools for studying such interactions. Objective-wise gradients expose differences in direction and magnitude that are unavailable from the summed gradient alone. Gradient aggregation methods use this information to construct an update, whereas scalarization methods combine objective values into a scalar function. Both are approaches to MOO. The distinction examined here is principally between fixed linear scalarization (LS) and aggregation based on objective-wise gradient information. Non-linear scalarization is also relevant and is included in our cell segmentation experiments.

\subsection{Background and Scope}
Gradient-based MOO has a substantial history, including the multiple gradient descent algorithm (MGDA)~\cite{desideri2012mutiple} and its application to multi-task neural networks~\cite{sener_2018}. Subsequent methods address gradient interference, alignment, and trade-offs through different update rules~\cite{yu2020gradient,liu2021conflict,senushkin2023independent,quinton2024jacobian}. Smooth Tchebycheff scalarization (STCH) provides a complementary approach based on a differentiable nonlinear scalar objective~\cite{lin_smooth_2024}. Applications also extend beyond conventional multi-task prediction, including multi-objective policy learning~\cite{parisi2014policy} and language-model alignment~\cite{MolmaPaperYage2025,zhong2024panacea,lee2024parrot}. Building on this literature, we investigate how established optimization tools behave across different objective decompositions within single-task model learning.

By \emph{single task}, we mean one prediction, generation, or decision problem, such as image classification, image reconstruction and generation, or molecule modification. A single task can nonetheless be trained with several loss terms and evaluated with several metrics. We examine two types of decomposition. The first separates the terms of a composite loss or reward, exposing interactions among training criteria. The second splits an empirical classification loss into class-conditional risks, exposing interactions among data groups. Neither decomposition captures every source of gradient conflict. In particular, within-class heterogeneity, changing environments, and conflicts over time require additional modeling choices that are outside the experiments reported here.

The presence of several terms does not make gradient aggregation preferable by itself. A scalar objective may already encode the intended utility, and a temporary increase in one component can be acceptable. Moreover, aggregation retains the choices about objective scaling, preference weights, normalization, and optimizer settings. Its additional computation is justified only if the resulting model offers useful performance or trade-off improvements. We therefore evaluate individual trained solutions and their task-specific metrics. We do not claim to recover a complete Pareto front or to establish global Pareto optimality for the trained networks because multiple training costs on GPUs are sequentially or even simultaneously infeasible.

\subsection{Research Questions and Contributions}
This study addresses three questions: (1) when do the evaluated multi-objective formulations improve on fixed weighted-sum training; (2) how do these outcomes depend on the aggregation method, objective normalization, and decomposition; and (3) what practical limitations affect the interpretation and use of the results? We examine long-tailed classification with 10 or 100 class-wise objectives, cell segmentation with six loss terms, generative models with two to five objectives, and molecular optimization with three reward components. Each application is presented with its formulation and empirical results, followed by a synthesis in \cref{sec:discussion}.

Our contributions are threefold. First, we formulate these single-task learning settings in a common objective-wise gradient notation and compare established MOO methods with their scalarized baselines. Second, we document both improvements and failures across the selected applications, distinguishing the effects of objective construction from the incremental effects of aggregation where the comparisons allow. Third, we derive practical considerations from these results, including normalization sensitivity, limits of gradient diagnostics, and measured runtime overhead. Overall, these contributions characterize the opportunities and limitations of applying established MOO algorithms across the examined single-task settings.

\section{Multi-Objective Formulation and Optimization}
\label{sec_moo}
Let $\{\mathcal{L}_1, \mathcal{L}_2, \ldots, \mathcal{L}_M\}$ denote our $M$ \textit{loss} functions. For each loss $\mathcal{L}_m$, we define the preference-weighted gradient $\boldsymbol{g}_m = \lambda_m \nabla_\theta \mathcal{L}_m$ with respect to the parameters $\theta$. We organize these gradients into a weighted Jacobian $\bm{J} \in \mathbb{R}^{M \times N}$, where $N=|\theta|$ is the number of model parameters and each row corresponds to one weighted objective gradient:
\begin{equation}
    \bm{J}
    =
    \underbrace{\begin{bmatrix}
    \boldsymbol{g}_1^\top \\
    \boldsymbol{g}_2^\top \\
    \vdots \\
    \boldsymbol{g}_M^\top
    \end{bmatrix}}_{\substack{\text{weighted}\\\text{gradients}}}
    =
    \underbrace{\mathrm{diag}(\bm{\lambda})}_{\substack{\text{preference}\\\text{weights}}}
    \underbrace{\begin{bmatrix}
    (\nabla_\theta \mathcal{L}_1)^\top \\
    (\nabla_\theta \mathcal{L}_2)^\top \\
    \vdots \\
    (\nabla_\theta \mathcal{L}_M)^\top
    \end{bmatrix}}_{\substack{\text{objective-wise}\\\text{gradients}}}.
\end{equation}
Here, $\lambda_m > 0$ is a user-specified preference and/or normalization weight applied before aggregation, and the $\lambda_m$ values do not need to have a sum of 1 (when the objectives are normalized already, the $\lambda_m$ values often have a sum of 1). They are distinct from the coefficients $\boldsymbol{\alpha}$ that some aggregation methods compute during training. For example, MGDA restricts $\boldsymbol{\alpha}$ to the probability simplex, but no such restriction applies to $\bm{\lambda}$. 
Gradient aggregation methods $\mathcal{A}$ operate on this weighted gradient matrix to compute an aggregated update direction that reduces gradient conflict and approaches Pareto stationarity \cite{quinton2024jacobian}:
$
\boldsymbol{G}_{\mathcal{A}} = \mathcal{A}(\bm{J}).
$
For an unconstrained differentiable problem, a parameter vector is Pareto optimal if no feasible parameter vector improves one objective without worsening another. Pareto stationarity is the weaker first-order condition that no direction strictly decreases all objectives. With strictly positive $\lambda_m$, it is equivalent to $\bm{0}\in\mathrm{conv}\{\bm{g}_1,\ldots,\bm{g}_M\}$~\cite{desideri2012mutiple,sener_2018}. Positive scaling preserves Pareto dominance and this stationarity condition, but it can change an aggregation method's trajectory. In nonconvex neural networks, stationarity alone does not guarantee Pareto optimality. The methods in the following differ in how they choose a direction and in the guarantees available for that direction.

\paragraph{Linear Scalarization (LS)} The simplest approach computes a weighted sum of losses $\mathcal{L}_{\text{total}} = \sum_{m=1}^{M} \lambda_m \mathcal{L}_m$, yielding an aggregated gradient $\boldsymbol{G}_{\mathrm{LS}}  = \sum_{m=1}^{M} \boldsymbol{g}_m$. 
A global minimizer with strictly positive weights is Pareto optimal, but varying the weights cannot recover unsupported points on a nonconvex Pareto front~\cite{das1997closer,lin_smooth_2024}.
This limitation concerns global trade-off coverage since it does not imply that LS cannot converge to a Pareto stationary point or that another method will produce a better model. LS remains the principal baseline in this study.

\paragraph{Multiple Gradient Descent Algorithm (MGDA)} To address the limitations of linear scalarization, Sener and Koltun~\cite{sener_2018} introduced MGDA for multi-task learning, which finds the minimum-norm element in the convex hull of the task gradients by solving:
\begin{equation}
    \min_{\boldsymbol{\alpha} \in \Delta^{M-1}}
\left\|
\sum_{m=1}^{M} \alpha_m \boldsymbol{g}_m
\right\|_2^2
=
\min_{\boldsymbol{\alpha} \in \Delta^{M-1}}
\|\bm{J}^\top \boldsymbol{\alpha}\|_2^2,
\end{equation}
where $\boldsymbol{\alpha}$ contains the aggregation coefficients, $\Delta^{M-1}$ is the probability simplex, and $\bm{G}_{\mathrm{MGDA}}=\sum_m\alpha_m\bm{g}_m$. 
The resulting direction is guaranteed to be a common descent direction for the weighted objectives, or indicates a Pareto stationary point under the weighted formulation. 
It is worth noting that the stochastic gradients, approximate solvers, finite step sizes, and subsequent optimizer transformations affect their application in training. MGDA can be expressed as a weighted gradient sum, with coefficients selected using inter-objective gradient geometry rather than fixed in advance.

\paragraph{Aligned Multi-Task Learning (AMTL)} Senushkin \textit{et al.}~\cite{senushkin2023independent} proposed AMTL to reduce both gradient dominance and gradient conflict before aggregating objectives. Let the objective-wise gradients form the matrix $\boldsymbol{J}^\top=[\boldsymbol{g}_1,\ldots,\boldsymbol{g}_M]$. AMTL first computes a compact SVD on its rank-$r$ subspace, $\boldsymbol{J}^\top=\boldsymbol{U}_r\boldsymbol{\Sigma}_r\boldsymbol{V}_r^\top$, the alignment replaces the retained positive singular values by a common scale $s$: 
\begin{equation}
\begin{aligned}
    \widetilde{\boldsymbol{J}}^\top
    &=
    \boldsymbol{U}_r(s\boldsymbol{I}_r)\boldsymbol{V}_r^\top,\\
    s &\in
    \{\mathrm{min}_i (\sigma_i),\ \mathrm{median}_i(\sigma_i),\ \mathrm{RMSE}_i(\sigma_i)\}.
\end{aligned}
\end{equation}
The columns of $\widetilde{\boldsymbol{J}}^\top$ are the aligned gradients $\widetilde{\boldsymbol{g}}_m$. AMTL then forms the final update by summing the aligned gradients: $\boldsymbol{G}_{\mathcal{A}_{\text{AMTL}}}
=
\sum_{m=1}^{M} \widetilde{\boldsymbol{g}}_m .$
The aligned gradients solve a generalized Procrustes problem~\cite{Schomemann1966,senushkin2023independent}. When the columns are linearly independent, this transformation makes them orthogonal with equal norm. Orthogonality of the transformed gradients does not guarantee a common descent direction for the original objectives. We evaluate minimum, median, root-mean-square (RMSE) scaling variants in this work. A small retained singular value can make the minimum scale update very small. The effect of these choices can be assessed empirically from one case to another rather than being assumed uniformly.

\paragraph{Unconflicting Projection of Gradients (UPGrad)} Within the Jacobian Descent (JD) framework \cite{quinton2024jacobian}, UPGrad projects each objective-wise gradient onto the dual cone formed by the rows of the weighted Jacobian matrix $\bm{J}$, then averages the projected results. The dual cone induced by $\bm{J}$ is $\bm{J}^{*} = \{ \boldsymbol{x}\in\mathbb{R}^{N}: \bm{J}\boldsymbol{x} \ge \bm{0}\}$, where the inequality is applied elementwise. The aggregated gradient is then $\boldsymbol{G}_{\mathcal{A}_{\text{UPGrad}}}  = \frac{1}{M} \sum_{m=1}^{M} \mathrm{proj}_{\bm{J}^{*}}(\boldsymbol{g}_m)$, where $\mathrm{proj}_{\bm{J}^{*}}(\cdot)$ denotes projection onto the dual cone. 
With exact projections, the resulting common direction $\boldsymbol{G}_{\mathcal{A}_{\text{UPGrad}}} $ has non-negative inner products with the objective-wise gradients. This is a first-order property before any momentum or adaptive preconditioning is applied.

\paragraph{Smooth Tchebycheff Scalarization (STCH)} Lin \textit{et al.}~\cite{lin_smooth_2024} introduced a smooth approximation to the classical Tchebycheff scalarization~\cite{beckmann_relationship_1976}. Under suitable preferences and reference point conditions, Tchebycheff formulations can reach non-convex regions of the Pareto front that weighted sums miss. Given a preference vector $\bm{\lambda}$ and reference point $\boldsymbol{z}$, the classical Tchebycheff objective minimizes the largest weighted deviation $\max_m \lambda_m(\mathcal{L}_m(\bm{\theta})-z_m)$. STCH makes this maximum differentiable by replacing the hard maximum with a soft maximum:
\begin{equation}
    \mathcal{L}_{\mathrm{STCH}}(\bm{\theta};\bm{\lambda},\boldsymbol{z})
    =
    \mu \log
    \sum_{m=1}^{M}
    \exp\left(
        \frac{\lambda_m(\mathcal{L}_m(\bm{\theta})-z_m)}{\mu}
    \right),
\end{equation}
where $\mu>0$ controls how soft the maximum is. As $\mu \to 0$, the largest weighted deviation dominates, recovering the classical Tchebycheff objective. Larger $\mu$ spreads attention across more objectives, so the gradient combines objectives according to their weighted deviations from the reference point. STCH can be differentiated as a single scalar loss and does not require Jacobian matrix.

\paragraph{Other methods} It is worth mentioning several other gradient aggregation methods designed for multi-task learning. PCGrad \cite{yu2020gradient} projects conflicting gradients onto the normal plane of each other to reduce interference. CAGrad \cite{liu2021conflict} seeks a common descent direction that maximizes the minimum improvement across all tasks. IMTL-G \cite{liu2021towards} rescales gradients to achieve balanced optimization. GradDrop \cite{chen2020just} randomly masks gradient components to reduce conflicts. Usually, the aggregation methods are stateless, meaning that they aggregate the gradients based on the given Jacobian matrix. However, NashMTL \cite{navon2022multi} is a stateful gradient aggregation method that forms a bargaining game and computes the Nash equilibrium of objectives. These methods are single-solution solvers which makes them attractive for handling gradient conflicts for large networks in multi-objective settings. Our experiments cover a subset of the methods, not an exhaustive ranking of MOO algorithms.

\subsection{A Local Diagnostic for Objective Interference}
\label{sec:local_diagnostic}
Pairwise gradient conflict and an adverse scalarized update are related but distinct. The following elementary consequence of differentiability identifies when an LS step locally increases an individual loss.

\begin{proposition}[Gradient conflict induces first-order ascent under LS]
\label{prop:ls_ascent}
Let $\mathcal{L}_1,\ldots,\mathcal{L}_M$ be differentiable at $\bm{\theta}$, let $\lambda_m>0$ and define $\boldsymbol{g}_m=\lambda_m \nabla_{\bm{\theta}}\mathcal{L}_m(\bm{\theta})$, and $\bm{d}_{\mathrm{LS}}=\sum_m\bm{g}_m$. If $\bm{g}_i^\top\bm{d}_{\mathrm{LS}}<0$ for some $i$, then $\mathcal{L}_i(\bm{\theta}-\eta\bm{d}_{\mathrm{LS}})>\mathcal{L}_i(\bm{\theta})$ for all sufficiently small $\eta>0$. 
For any fixed direction $\bm{d}$, the first-order change of objective $m$ is nonpositive when $\bm{g}_m^\top\bm{d}\geq0$, and strictly negative when $\bm{g}_m^\top\bm{d}>0$.
\end{proposition}
\begin{proof}
For any fixed $\bm{d}$, differentiability returns:
\begin{equation}
\mathcal{L}_m(\bm{\theta}-\eta\bm{d})
=\mathcal{L}_m(\bm{\theta})-\frac{\eta}{\lambda_m}\bm{g}_m^\top\bm{d}+o(\eta).
\end{equation}
Since $\lambda_m>0$, a negative inner product makes the linear term strictly positive, which dominates the remainder for sufficiently small $\eta$. The remaining statements follow from the sign of the same linear term. When the inner product is zero, higher-order terms determine the actual loss change.
\end{proof}

This proposition gives a first-order boundary between fixed scalarization and dynamic gradient aggregation. If no objective has $\boldsymbol{g}_m^\top \boldsymbol{d}_{\mathrm{LS}}<0$, LS is already a common descent step to first order. If pairwise conflicts $\boldsymbol{g}_i^\top\boldsymbol{g}_j<0$ are sufficiently strong, the weighted sum above can become negative for some objective $\mathcal{L}_i$, so LS locally improves the scalarized loss while increasing an individual objective. Equivalently, conflict angles greater than $90^\circ$ (negative cosine similarity) can make fixed LS update directions incompatible when the non-convex Pareto front on objectives limits what any fixed linear scalarization can reach \cite{das1997closer}. MOO methods such as MGDA, AMTL, and UPGrad target this specific failure mode by searching for an update with non-negative inner product against each objective gradient, or by certifying that no such common descent direction exists near Pareto stationarity.

The relevant quantity is
\begin{equation}
\bm{g}_i^\top\bm{d}_{\mathrm{LS}}
=\underbrace{\|\bm{g}_i\|_2^2}_{\substack{\text{self-alignment}\\\geq\,0}}
+\underbrace{\sum_{j\ne i}\bm{g}_i^\top\bm{g}_j}_{\substack{\text{cross-objective}\\\text{interaction}}}.
\end{equation}
A negative pairwise inner product alone is insufficient to make this sum negative. Conversely, an adverse step on one component need not be undesirable if it improves the intended scalar utility. These diagnostics therefore identify local interference, not a sufficient condition for better validation performance with MOO. They also do not measure distance to a Pareto front. For momentum or adaptive optimizers, the actual parameter displacement can differ from $-\eta\bm{G}_{\mathcal{A}}$, and interpreting first-order behavior then requires evaluating the objective gradients against that displacement.

\section{Multi-Class Classification as Multi-Objective Optimization}
A multi-class classification task with $M$ classes is typically trained by empirical risk minimization (ERM) using cross-entropy (CE) \cite{bishop2006pattern}. 
For an input $\bm{x}$ with one-hot label vector $\bm{y}\in\{0,1\}^M$, the Multinoulli conditional distribution is
\begin{align}
p(\bm{y}\mid\bm{x}) = \prod_{m=1}^M \mu_m^{y_m},\label{eq_multinoulli}
\end{align}
where $\sum_{m=1}^M \mu_{m}=1$ and $0<\mu_{m}<1$. A neural network with parameters $\bm{\theta}$ and logits $f_m(\bm{x},\bm{\theta})$ computes these parameters with the softmax function
\begin{align}
\mu_m(\bm{x};\bm{\theta}) = \frac{\exp(f_m(\bm{x},\bm{\theta}))}{\sum_{m'=1}^M \exp(f_{m'}(\bm{x},\bm{\theta}))}.
\end{align}
Let $\hat{p}(\bm{x},\bm{y})$ denote the empirical distribution induced by the training set. Taking the negative logarithm of \cref{eq_multinoulli} and averaging over $\hat{p}$ gives the CE objective
\begin{align}
\mathcal{L}_{\mathrm{CE}}(\bm{\theta})
= \mathbb{E}_{(\bm{x},\bm{y})\sim \hat{p}}\Big[-\sum_{m=1}^M y_m\log \mu_m(\bm{x};\bm{\theta})\Big].
\end{align}
Since $\bm{y}$ is one-hot, only the term of the true class $y$ remains, where we use the scalar variable $y\in\{1,\cdots,M\}$ to represent the categorical representation of class labels corresponding to the one-hot vector $\bm{y}$. For a mini-batch $\mathcal{B}$, the stochastic ERM objective is therefore
\begin{align}
\mathcal{L}_{\mathrm{ERM}}(\bm{\theta};\mathcal{B})=\hat{\mathbb{E}}_{(\bm{x},y)\in\mathcal{B}}\Big[-\log \mu_y(\bm{x};\bm{\theta})\Big].
\end{align}
ERM averages sample risks from different classes in a mini-batch into a single scalar objective. Standard cross-entropy gives each example equal weight, so each class contributes to the batch loss in proportion to its share of the batch. The sampling scheme determines the expected class proportions. Even with a balanced sampling mechanism, the summed gradient alone does not identify inter-class conflicts. Such conflicts can arise in multi-class networks as the class-wise risks are optimized over shared parameters, and softmax normalization couples the class probabilities in the classification head.

We decompose the empirical risk by true class. Let $\pi_m=\Pr_{\hat{p}}(y=m)$ denote the probability of drawing class $m$ under the chosen sampling scheme, and let $\mathcal{B}_m=\{\bm{x}:(\bm{x},y)\in\mathcal{B},\,y=m\}$ be the subset of the mini-batch belonging to class $m$. For each class present in the sampled mini-batch, its empirical conditional loss is as follows:
\begin{align}
\mathcal{L}^m_{\mathrm{CE}}(\bm{\theta};\mathcal{B})
&= \hat{\mathbb{E}}_{\bm{x}\in\mathcal{B}_m}\Big[-\log \mu_m(\bm{x};\bm{\theta})\Big].
\end{align}
This decomposition exposes pairwise inter-class gradient conflicts, i.e., $\nabla_\theta \mathcal{L}^i_{\mathrm{CE}}{}^\top \nabla_\theta \mathcal{L}^j_{\mathrm{CE}} < 0$. We use these class-wise losses as the objectives for gradient aggregation.

\textbf{Class-wise Risk Minimization (CWRM).}
The vector of class-wise risks defines the CWRM formulation as follows:
\begin{align}
\min_{\bm{\theta}}\ \bm{\mathcal{L}}_{\mathrm{CWRM}}(\bm{\theta};\mathcal{B}) 
= \big[\mathcal{L}^1_{\mathrm{CE}}(\bm{\theta};\mathcal{B}),\ldots,\mathcal{L}^M_{\mathrm{CE}}(\bm{\theta};\mathcal{B})\big]^\top.
\end{align}
For a sampled mini-batch, let $\mathcal{I}_{\mathcal{B}}=\{m:|\mathcal{B}_m|>0\}$ denote its represented classes. The exact decomposition of the sample averaging ERM loss is as follows:
\begin{equation}
\label{eq:erm_decomposition}
\mathcal{L}_{\mathrm{ERM}}(\bm{\theta};\mathcal{B})=
\sum_{m\in\mathcal{I}_{\mathcal{B}}}
\underbrace{\frac{|\mathcal{B}_m|}{|\mathcal{B}|}}_{\substack{\text{weight on the risk}\\\text{for class }m}}
\underbrace{\mathcal{L}^m_{\mathrm{CE}}(\bm{\theta};\mathcal{B})}_{\text{class-wise risk}}.
\end{equation}
Thus, decomposing the loss only leaves ERM unchanged when the same coefficients and sampling are retained. Equal class weighting instead emphasizes class-conditional risks that are separate from empirical class frequency in the training dataset, as do related class-balancing approaches~\cite{cui2019class}. In our experiments, CWRM uses class-balanced sampling, whereas ERM samples uniformly from the imbalanced training set  (see Appendix~\ref{S-app:longtail} in the supplementary material). CWRM with LS (CWRM+LS), therefore, provides the relevant scalarized comparison for assessing the incremental effect of gradient aggregation. Its comparison with ERM would reflect the change in sampling and class weighting.

Everson and Fieldsend~\cite{everson2006multiclass} studied trade-offs among class-specific errors in multi-class ROC analysis. They approximated the Pareto front of pairwise misclassification rates and also examined a reduced formulation with one error-rate objective per class. On a fixed dataset, these empirical error rates are piecewise constant in the model scores. CWRM defines a separate objective for each class by averaging a chosen loss over examples from that class. In this study, we compute the gradient of the mean cross-entropy loss for each class and aggregate these gradients to update the shared model parameters. This formulation lets us examine how gradient aggregation affects learning under class imbalance.

\begin{table}[!t]
    \centering
    \caption{\textbf{Long-tailed classification.} Top-1 accuracy (\%) on CIFAR10/100-LT test dataset trained under different imbalance factors. The results are averaged over 3 independent seeds, and the best and second-best results for each experiment setting are shown in bold and underlined, respectively. $\dagger$: $T=250$, $\epsilon = 0.001$.
    $\ddagger$: $T=10$, $\epsilon = 0.1$.}
    \label{tab:cifar_lt}
    \resizebox{\columnwidth}{!}{%
    \begin{tabular}{lcccccc}
    \toprule
    \textbf{Method} & \multicolumn{3}{c}{\textbf{CIFAR10-LT}} & \multicolumn{3}{c}{\textbf{CIFAR100-LT}} \\
    \cmidrule(lr){2-4}\cmidrule(lr){5-7}
    \textbf{Imbalance Factor (IF)} & 50 & 100 & 200 & 50 & 100 & 200 \\
    \midrule
    ERM & 77.53 & 71.97 & 66.07 & 44.47 & 38.96 & 35.76 \\
    \midrule
    CWRM + LS & 83.14 & {78.65} & 71.30 & 47.93 & 42.46 & {37.99} \\
    CWRM + AMTL-min & {83.20} & 78.46 & {71.78} & 44.71 & 36.85 & 30.57 \\
    CWRM + AMTL-RMSE & {83.38} & 78.41 & 71.60 & 48.29 & \underline{42.71} & \underline{38.58} \\
    CWRM + AMTL-median & \underline{83.71} & 78.51 & \underline{72.05} & \underline{48.30} & 42.69 & 38.53 \\
    CWRM + UPGrad & 82.71 & {79.03} & 70.37 & \textbf{48.46} & \textbf{42.88} & \textbf{38.73}\\
    CWRM + MGDA $\dagger$ & 83.63 & \textbf{79.51} & 71.93 & 37.41 & 29.18 & 20.93\\
    CWRM + MGDA $\ddagger$ & \textbf{83.97} & \underline{79.13} & \textbf{73.49} & 38.73 & {31.78} & {37.50}\\
    CWRM + MGDA-GN $\dagger$ & 83.15 & 78.57 & 72.04 & {47.43} & {42.39} & {34.13}\\
    CWRM + MGDA-LN $\dagger$ & 83.69 & 78.63 & 71.40 & \underline{48.30} & {42.70} & {37.85}\\
    CWRM + MGDA-LGN $\dagger$ & 83.34 & 78.34 & 71.24 & {46.16} & {40.43} & {17.77}\\
    \bottomrule
    \end{tabular}
    }
\end{table}

\Cref{tab:cifar_lt} shows that CWRM+LS improves over ERM in all six settings. Gradient aggregation yields additional gains in selected configurations, but its effect depends on the dataset and imbalance factor. On CIFAR10-LT, the best MGDA configurations improve on CWRM+LS by 0.83, 0.86, and 2.19 percentage points at imbalance factors 50, 100, and 200, respectively. On CIFAR100-LT, UPGrad has the highest reported mean accuracy at all three imbalance factors, exceeding CWRM+LS by 0.53, 0.42, and 0.74 percentage points. For example, at IF=100, the change from ERM to CWRM+LS is 38.96\% to 42.46\%, while UPGrad further increases accuracy to 42.88\%. This separates the larger gain associated with class-balanced CWRM from the smaller incremental gain of aggregation.

The failures are also informative. AMTL-min and several MGDA configurations substantially underperform CWRM+LS on CIFAR100-LT, despite competitive results for some of these methods on CIFAR10-LT. AMTL-median and AMTL-RMSE are more consistent in the evaluated 100-class settings. These results motivate checking normalization, singular-value scaling, and solver settings as the number of objectives increases; the simultaneous changes in dataset and class count prevent attributing the differences to objective count alone.

\begin{figure*}[!t]
    \centering
    \includegraphics[width=0.78\textwidth]{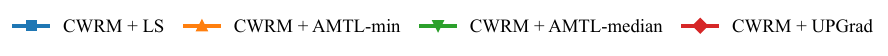}
    \vspace{0.3em}

    \subfloat[Gradient similarity\label{fig:cwrm_gradient_similarity}]{\includegraphics[width=0.32\textwidth]{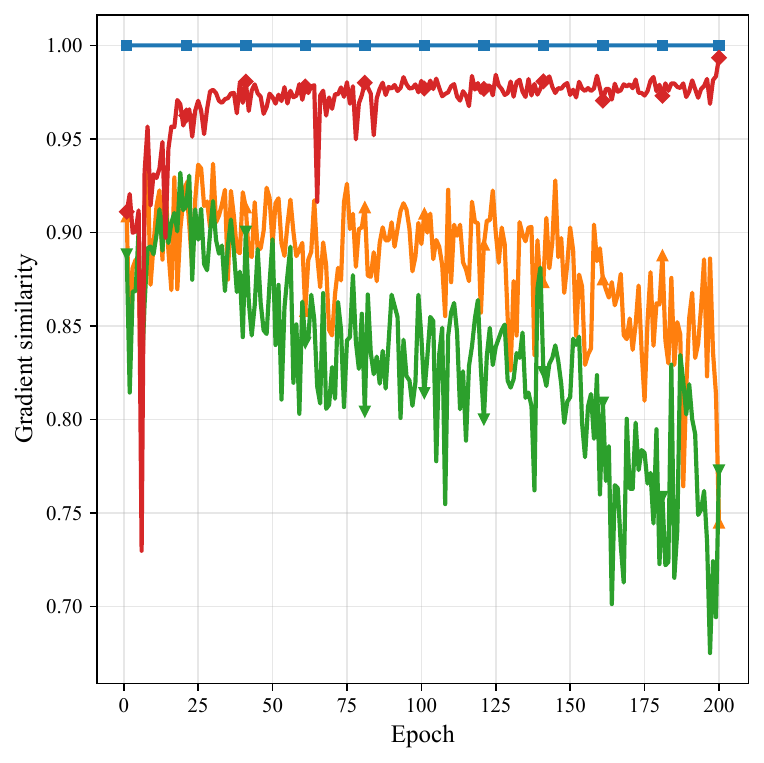}}
    \hfill
    \subfloat[Gradient inner product\label{fig:cwrm_pairwise_inner}]{\includegraphics[width=0.32\textwidth]{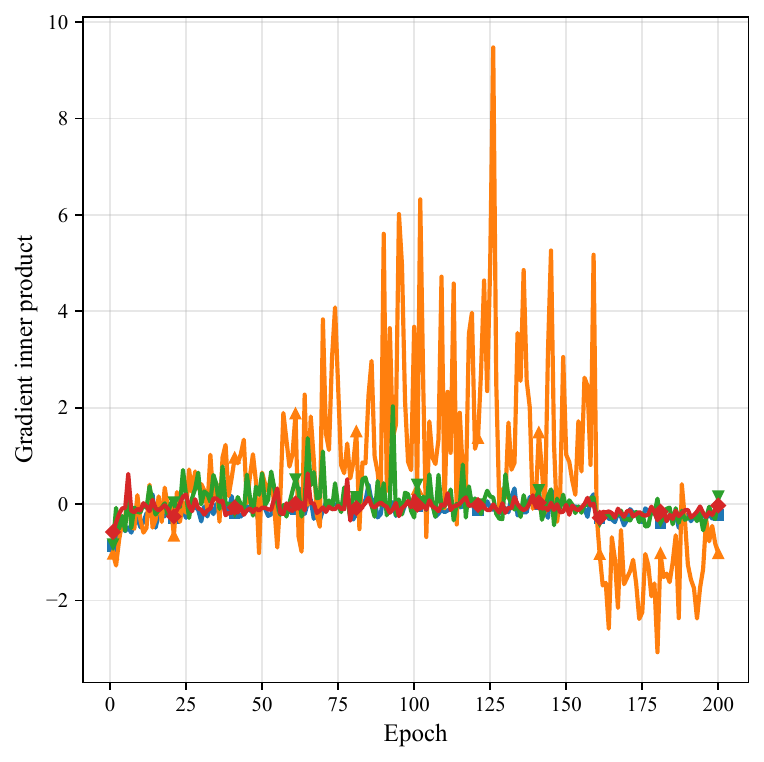}}
    \hfill
    \subfloat[Gradient conflict energy\label{fig:cwrm_conflict_energy}]{\includegraphics[width=0.32\textwidth]{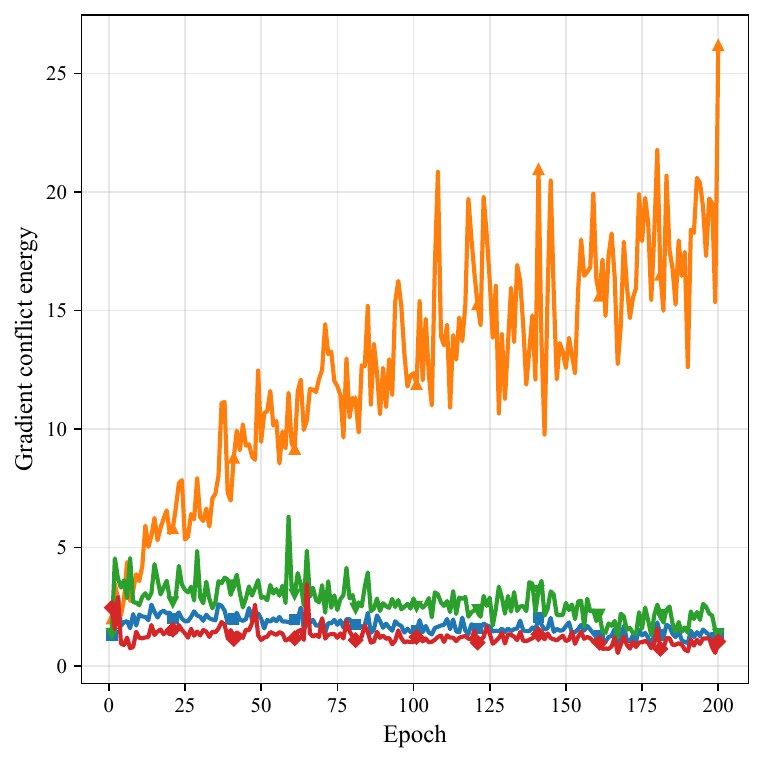}}
    \caption{\textbf{CWRM gradient diagnostics on CIFAR100-LT with imbalance factor of 50.} (a) Cosine similarity between each aggregated update and the CWRM+LS direction. (b) Mean pairwise inner product of the class-wise gradients. (c) Mean magnitude of the negative pairwise inner products. The metrics are defined in Appendix~\ref{S-app:gradient_diagnostics} in the supplementary material.}
    \label{fig:cwrm_gradient_diagnostics}
\end{figure*}

\Cref{fig:cwrm_gradient_diagnostics} tracks the class-wise gradient geometry during training on CIFAR100-LT with IF=50. UPGrad stays close to the LS direction throughout training, with a similarity of about 0.97--0.98, whereas both AMTL variants rotate the update further away from it. The conflict energy among class-wise gradients remains low for LS, UPGrad, and AMTL-median, but grows steadily for AMTL-min, which also has the lowest accuracy of the plotted methods in this setting (44.71\% versus 47.93\% for CWRM+LS).

\begin{figure*}[!t]
    \centering
    \subfloat[IF = 50\label{fig:top1_cifar10_if50}]{\includegraphics[width=0.29\textwidth]{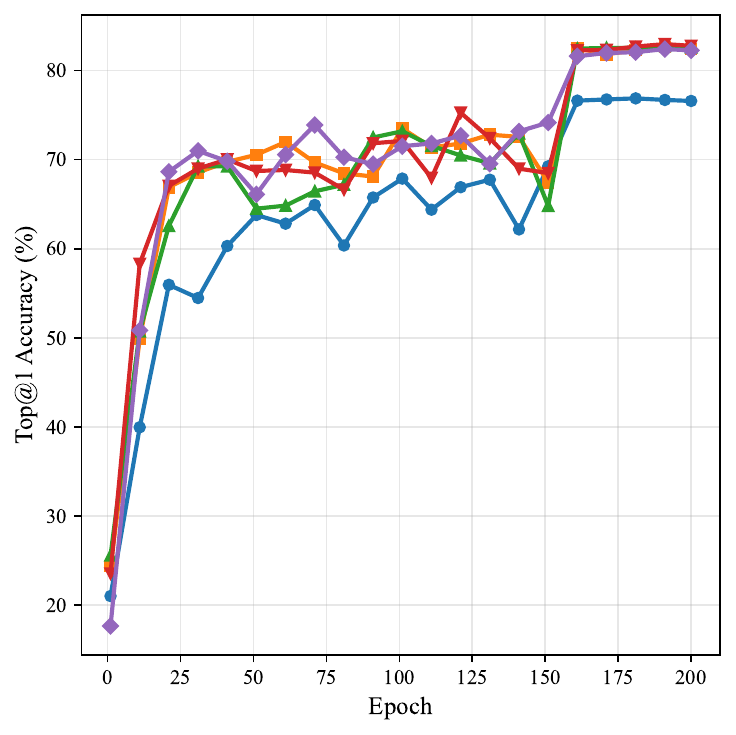}}
    \hfill
    \subfloat[IF = 100\label{fig:top1_cifar10_if100}]{\includegraphics[width=0.29\textwidth]{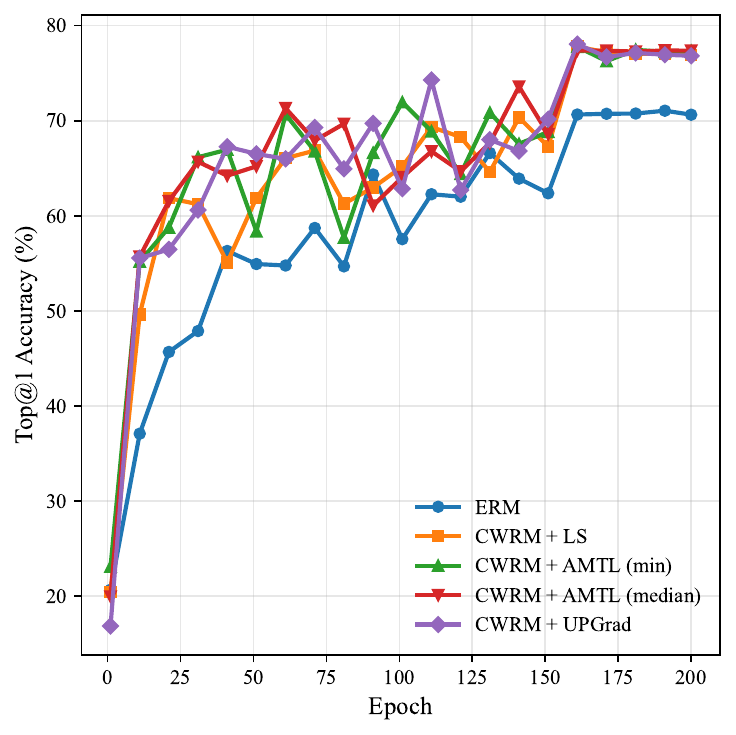}}
    \hfill
    \subfloat[IF = 200\label{fig:top1_cifar10_if200}]{\includegraphics[width=0.29\textwidth]{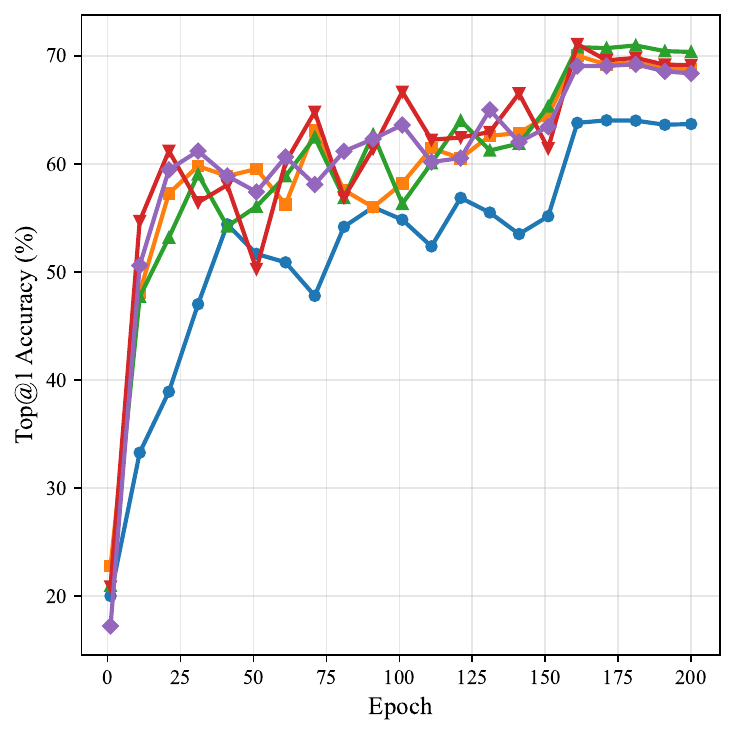}}
    \caption{Top-1 test accuracy during training on CIFAR10-LT at different imbalance factors.}
    \label{fig:top1_cifar10}
\end{figure*}

\begin{figure*}[!t]
    \centering
    \subfloat[IF = 50\label{fig:top1_cifar100_if50}]{\includegraphics[width=0.29\textwidth]{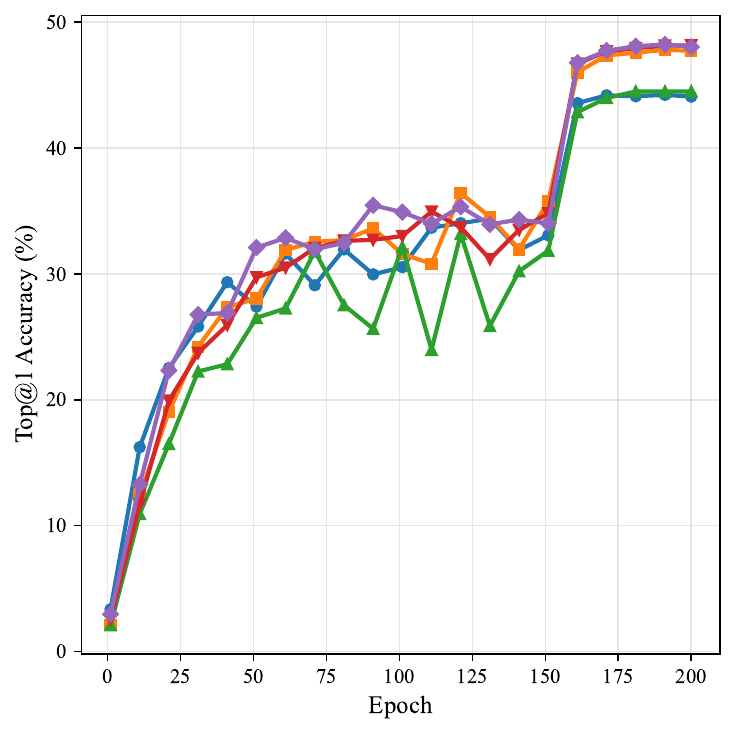}}
    \hfill
    \subfloat[IF = 100\label{fig:top1_cifar100_if100}]{\includegraphics[width=0.29\textwidth]{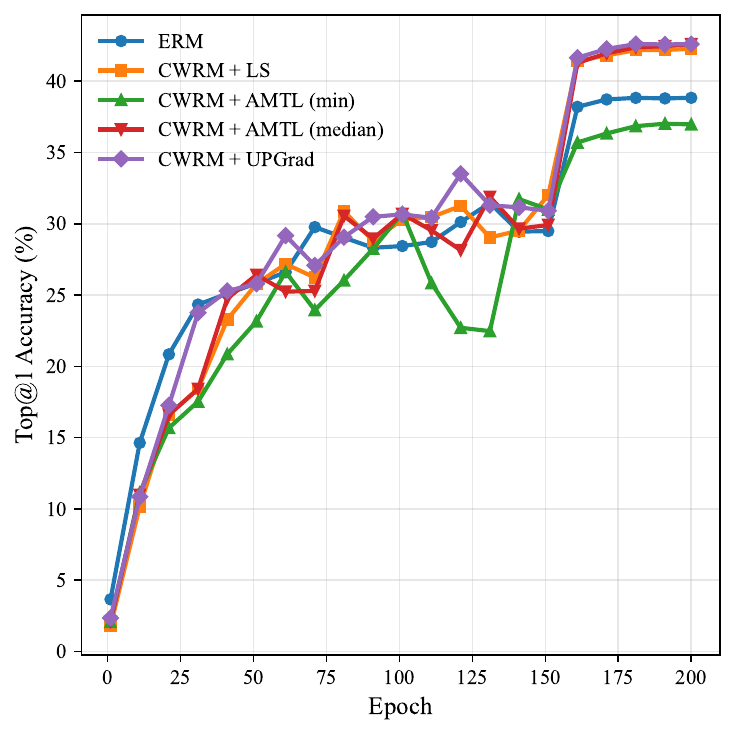}}
    \hfill
    \subfloat[IF = 200\label{fig:top1_cifar100_if200}]{\includegraphics[width=0.29\textwidth]{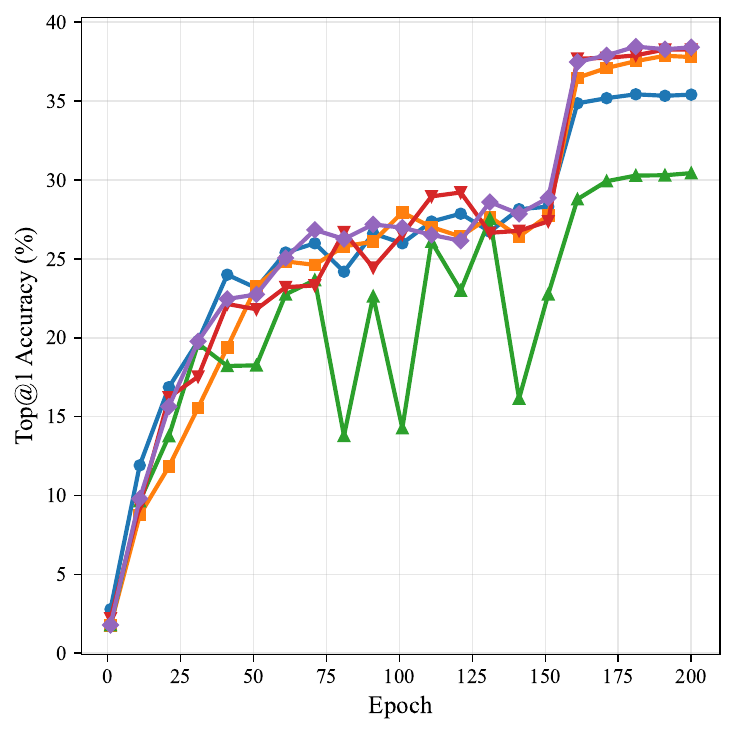}}
    \caption{Top-1 test accuracy during training on CIFAR100-LT at different imbalance factors.}
    \label{fig:top1_cifar100}
\end{figure*}

\begin{figure*}[!t]
    \centering
    \subfloat[IF = 50\label{fig:radar_cifar10_if50}]{\includegraphics[width=0.32\textwidth]{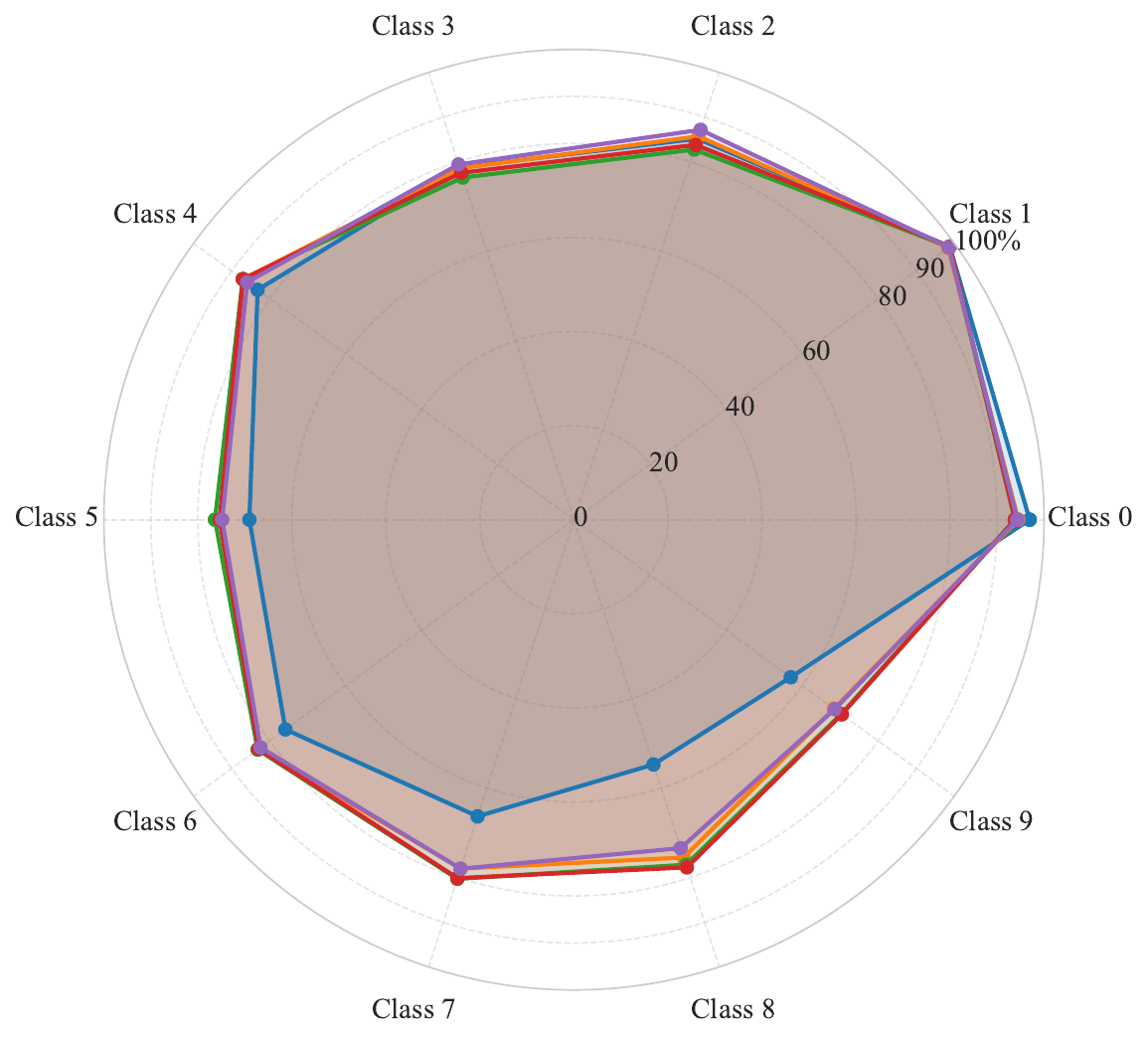}}
    \hfill
    \subfloat[IF = 100\label{fig:radar_cifar10_if100}]{\includegraphics[width=0.32\textwidth]{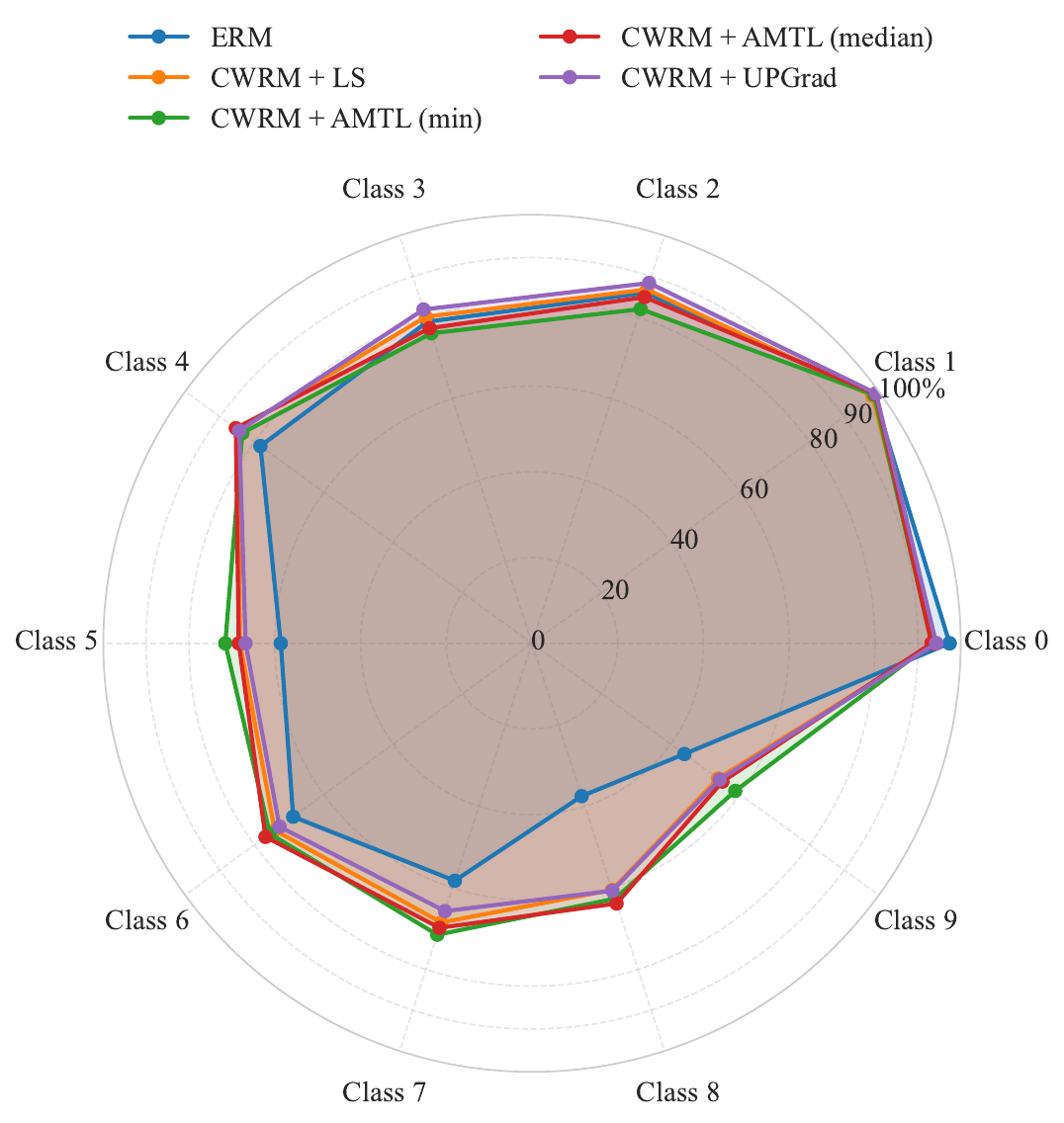}}
    \hfill
    \subfloat[IF = 200\label{fig:radar_cifar10_if200}]{\includegraphics[width=0.32\textwidth]{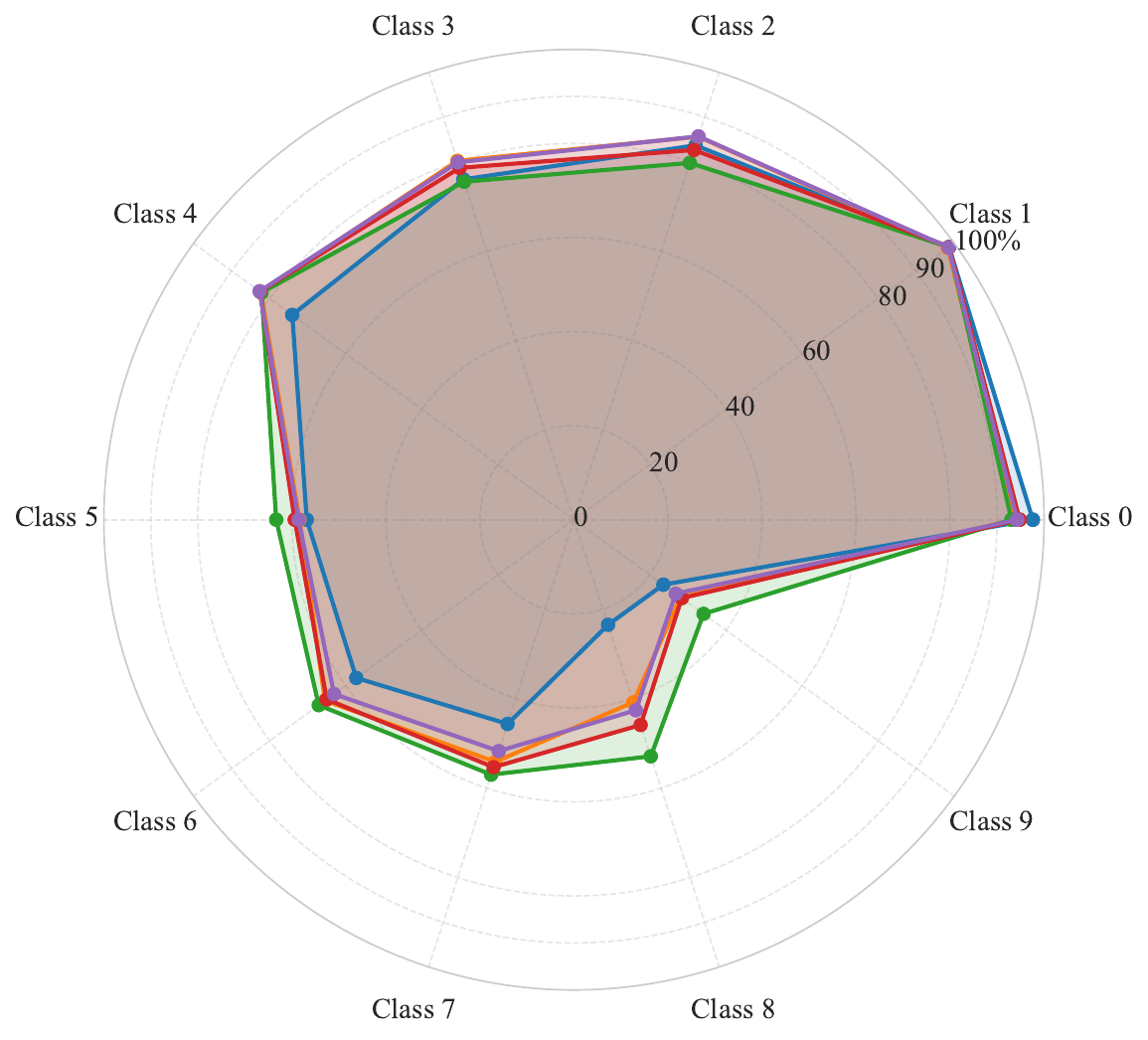}}
    \caption{Per-class test accuracy on CIFAR10-LT at different imbalance factors. Class indices are ordered from the most frequent (head) class 0 to the least frequent (tail) class 9.}
    \label{fig:radar_cifar10}
\end{figure*}

\begin{figure*}[!t]
    \centering
    \subfloat[IF = 50\label{fig:radar_cifar100_if50}]{\includegraphics[width=0.32\textwidth]{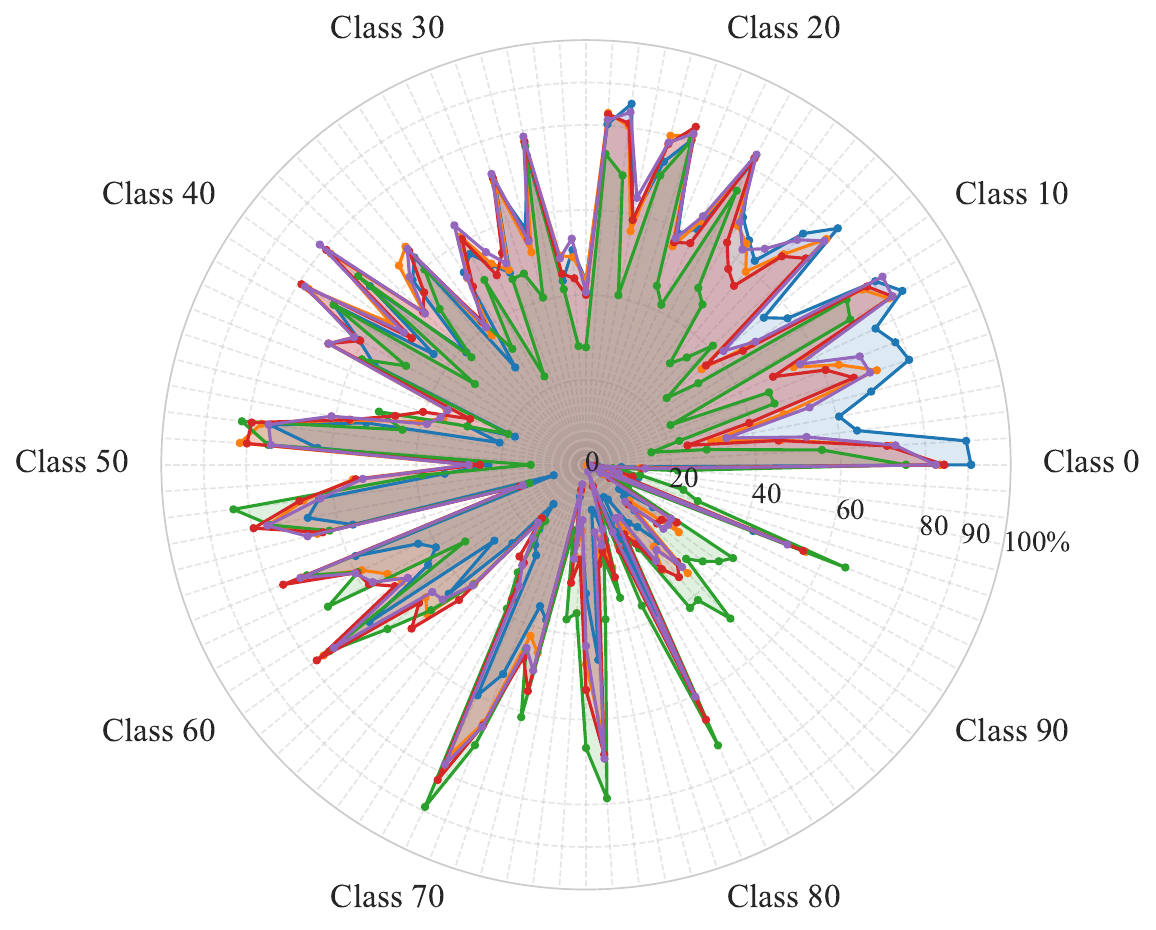}}
    \hfill
    \subfloat[IF = 100\label{fig:radar_cifar100_if100}]{\includegraphics[width=0.32\textwidth]{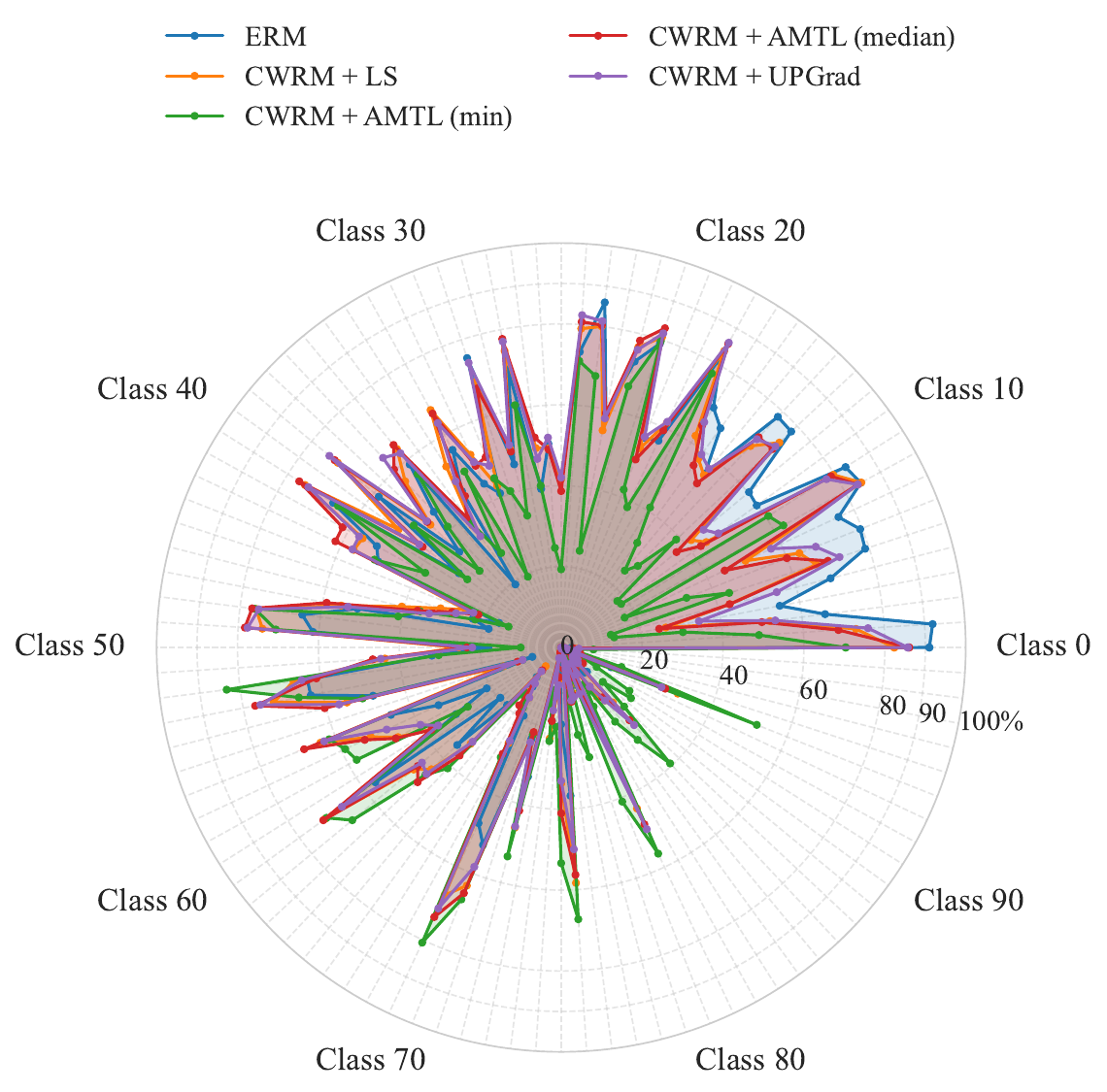}}
    \hfill
    \subfloat[IF = 200\label{fig:radar_cifar100_if200}]{\includegraphics[width=0.32\textwidth]{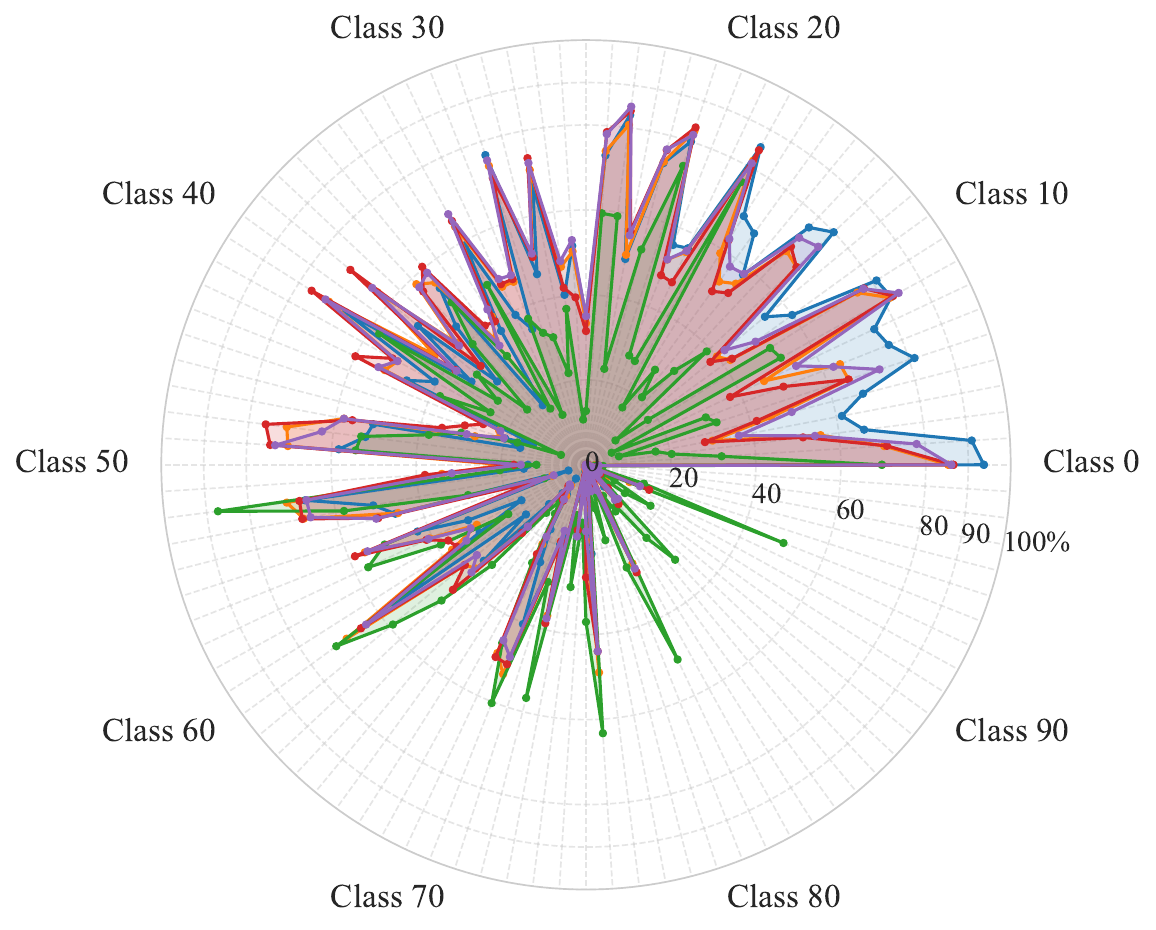}}
    \caption{Per-class test accuracy on CIFAR100-LT at different imbalance factors. Class indices are ordered from the most frequent (head) class 0 to the least frequent (tail) class 99.}
    \label{fig:radar_cifar100}
\end{figure*}

\Cref{fig:top1_cifar10,fig:top1_cifar100} show top-1 test accuracy during training for ERM and the CWRM variants. Before the learning-rate decay at epoch 160, accuracy fluctuates substantially and the curves cross frequently, so method rankings in this phase are not informative. The separation between ERM and CWRM appears after the decay and persists to the end of training. On CIFAR100-LT, AMTL-min finishes below ERM, consistent with \cref{tab:cifar_lt}. \Cref{fig:radar_cifar10,fig:radar_cifar100} show where these gains arise. On CIFAR10-LT, ERM matches or exceeds the CWRM variants on the head classes but has lower accuracy on the tail classes, whereas the CWRM variants largely overlap. On CIFAR100-LT (\cref{fig:radar_cifar100}), ERM again leads on the head classes, and accuracy is low for all methods on the tail classes. AMTL-min reaches higher accuracy than the other methods on several tail classes, but its overall accuracy falls below that of ERM at IF=100 and IF=200 (\cref{tab:cifar_lt}). The per-class view therefore supports the conclusion that class-balanced decomposition accounts for most of the improvement over ERM, with smaller differences among aggregation methods. Appendix~\ref{S-app:classification_details} in the supplementary material provides the training protocol.

\section{Self-Supervised Cell Segmentation}\label{sec:segmentation}
Traditionally, cell segmentation is treated as a supervised learning task (i.e., pixel classification), where ground truth is provided as a binary matrix of the same size as the input image \cite{Olaf2015}. Recently, supervised learning has become practically impossible for the recently developed spatial transcriptomics (ST) technology \cite{stahl_visualization_2016}, which enables the visualization and analysis of single-cell sequencing data in the spatial context. Spatial transcriptomics data are characterized by ultra-high-resolution images and multimodality such as nuclear stains images, RNA transcripts in a spatial density map, and cell type information. Segmentation of cells from spatial transcriptomics data is essential for subsequent analysis. 
However, manual annotation of cell boundaries is impractical due to the sheer number of cells and the heterogeneity of cell types and morphology~\cite{stahl_visualization_2016,fu_bidcell_2024}. 
Self-supervised methods such as BIDCell therefore combine biologically informed losses to train a segmentation network~\cite{fu_bidcell_2024}.
Using the BIDCell U-Net 3+ architecture, we compare LS with MGDA, AMTL, UPGrad, and STCH for the following six-objective formulation:
\begin{equation}
    \min_{\bm{\theta}}\ \bm{\mathcal{L}}_{\mathrm{SSCS}}(\bm{\theta}) =
    \big[\mathcal{L}_{\mathrm{ne}}, \mathcal{L}_{\mathrm{cc}}, \mathcal{L}_{\mathrm{os}}, \mathcal{L}_{\mathrm{mu}}, \mathcal{L}_{\mathrm{pos}}, \mathcal{L}_{\mathrm{neg}}\big]^\top,
    \label{eq:sscs}
\end{equation}
where $\mathcal{L}_{\mathrm{ne}}$ is nuclei encapsulation loss, $\mathcal{L}_{\mathrm{cc}}$ is cell calling loss, $\mathcal{L}_{\mathrm{os}}$ is over-segmentation loss, $\mathcal{L}_{\mathrm{mu}}$ is multiple assignment loss, $\mathcal{L}_{\mathrm{pos}}$ and $\mathcal{L}_{\mathrm{neg}}$ are positive and negative marker losses.
The loss definitions and experimental protocol are given in Appendix~\ref{S-sec_exp_moseg} in the supplementary material. The implementation replaces BIDCell's overlap loss with a multi-assignment loss, so the comparison concerns this adapted training formulation.

\begin{figure*}[!t]
    \centering
    \subfloat[Normalized spatial measures.\label{fig:moseg_radar_a}]{\includegraphics[width=0.35\textwidth]{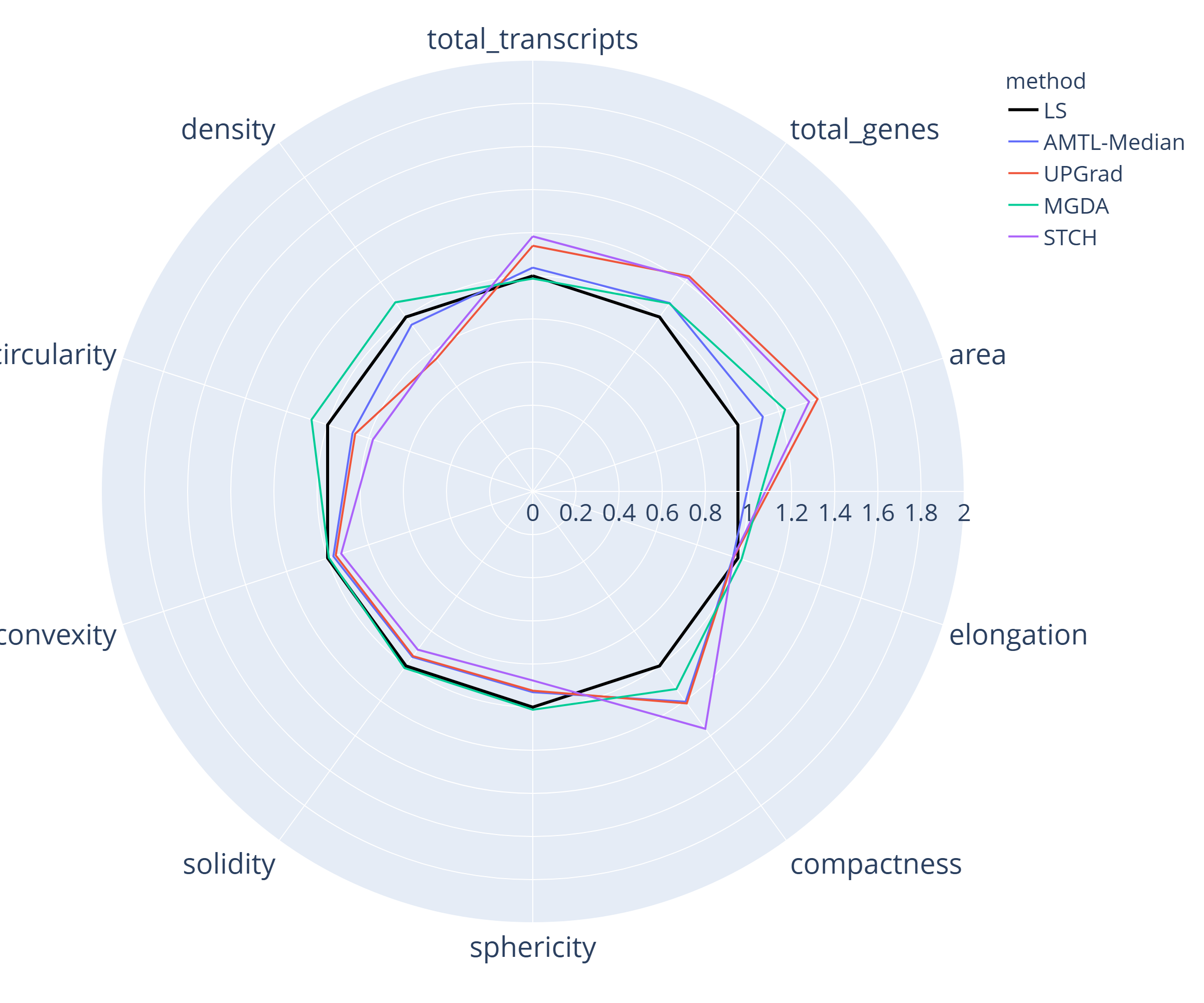}}
    \hfil
    \subfloat[Test sample segmentation outputs.\label{fig:moseg_segm_b}]{\includegraphics[width=0.61\textwidth]{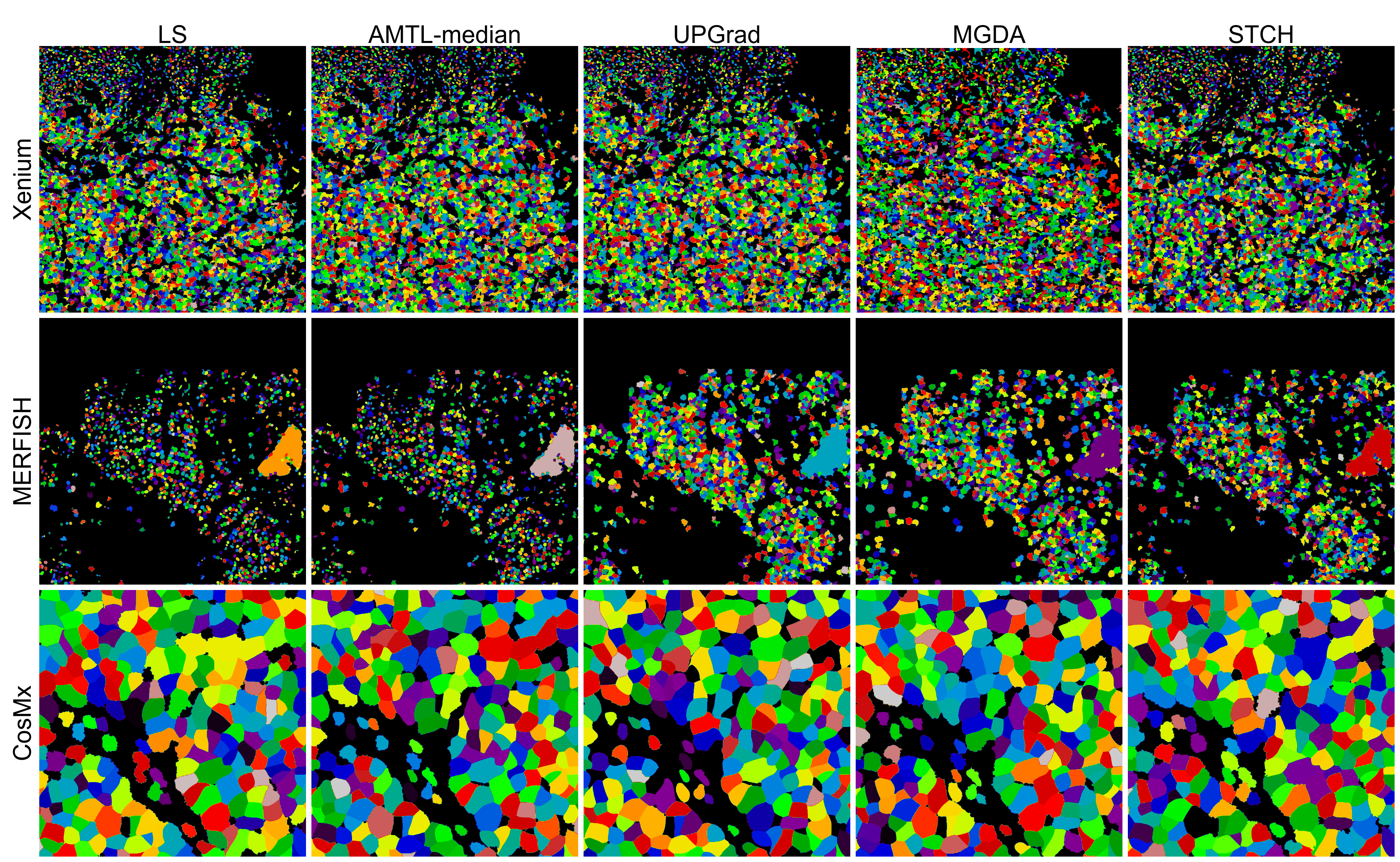}}
    \caption{\textbf{Cell segmentation}. (a) Normalized spatial measures across three ST platforms (Xenium, MERFISH, and CosMx). The LS baseline is shown as the unit circle ($=1.0$) and the values greater than 1.0 indicate larger values of that measure, not necessarily better segmentation. (b) Corresponding test sample segmentation outputs for LS and four MOO methods.}
    \label{fig:moseg_radar_segm}
\end{figure*}

The comparison in the radar chart is shown in \Cref{fig:moseg_radar_a}, where each axis represents a spatial morphological metric normalized against the LS baseline ($=1.0$). 
The LS method produces geometrically regular but biologically conservative segmentation cells are compact and round, yet under-assigned in transcript coverage. In contrast, MOO methods explicitly balance conflicting objectives (e.g., transcript assignment vs.\ shape regularity vs.\ overlap minimization), enabling exploration of different trade-offs. Among the MOO methods, AMTL achieves a well-balanced trade-off, expanding cell territories to capture more transcripts and genes while incurring only moderate reductions in shape regularity metrics such as circularity and compactness. UPGrad follows a similar trend with slightly more aggressive expansion. MGDA produces the most irregular cell morphologies, while it achieves large cell areas and high transcript density, it substantially sacrifices shape regularity, yielding cells with markedly lower circularity, convexity, solidity, and compactness than LS. STCH occupies a middle ground, deviating from LS on area and density metrics while retaining relatively more of the baseline's morphological regularity compared to AMTL and MGDA.
The corresponding segmentation examples in \Cref{fig:moseg_segm_b} qualitatively corroborate these quantitative findings on all three ST platforms (Xenium, MERFISH, and CosMx). MOO methods generally produce larger cells that fill more tissue area than LS, consistent with the increased cell area and gene capture seen in the radar chart. The visual differences are most pronounced for MERFISH and CosMx, where LS under-segments cells while AMTL and UPGrad yield more biologically plausible boundaries. No single method dominates all metrics simultaneously, which justifies the MOO formulation, as it allows practitioners to select the trade-off most appropriate for their downstream biological analysis.

\section{Variational Generative Models With Multi-Objective Optimization}\label{sec:generative}

Variational Autoencoders (VAEs)~\cite{Kingma2014} combine reconstruction and latent regularization within one generative modeling task. With encoder $q_\phi(\bm{z}\mid\bm{x})$, decoder $p_\theta(\bm{x}\mid\bm{z})$, and prior $p(\bm{z})$, define the following:
\begin{align}
\mathcal{L}_{\mathrm{recon}}(\theta,\phi)
&=-\mathbb{E}_{q_\phi(\bm{z}\mid\bm{x})}\!\left[\log p_\theta(\bm{x}\mid\bm{z})\right],\\
\mathcal{L}_{\mathrm{kld}}(\phi)
&=D_{\mathrm{KL}}\!\left(q_\phi(\bm{z}\mid\bm{x})\|p(\bm{z})\right).
\end{align}
The usual negative evidence lower bound (ELBO) is $\mathcal{L}_{\mathrm{recon}}+\mathcal{L}_{\mathrm{kld}}$. A coefficient $\beta>0$ yields the weighted sum training criterion $\mathcal{L}_{\mathrm{recon}}+\beta\mathcal{L}_{\mathrm{kld}}$, with $\beta=1$ recovering the negative ELBO. Our experiments use the preference-weighted vector denoted as follows:
\begin{equation}
\min_{\theta,\phi}\;
\big[\lambda_{\mathrm{recon}}\mathcal{L}_{\mathrm{recon}}(\theta,\phi),\ \lambda_{\mathrm{kld}}\mathcal{L}_{\mathrm{kld}}(\phi)\big]^\top.
\end{equation}
The weights $\lambda_{\mathrm{recon}}$ and $\lambda_{\mathrm{kld}}$ still set the relative scale of the two objectives, so treating them as separate objectives does not remove the need to choose these weights. We compute both objectives' gradients in the common parameter space $(\theta,\phi)$, with zero decoder components for the KL term, and apply the following:
\begin{equation}
\bm{G}_{\mathrm{VAE}}=\mathcal{A}\!\left([\bm{g}_{\mathrm{recon}}^\top;\bm{g}_{\mathrm{kld}}^\top]\right).
\end{equation}
This leads to changing the update rule while retaining the encoder, decoder, and particular losses. The comparison tests show how aggregation handles the reconstruction and regularization trade-off under the preference weights $\lambda_{\mathrm{recon}}$ and $\lambda_{\mathrm{kld}}$.

\begin{table}[!t]
    \centering
    \caption{\textbf{VAE reconstruction on CelebA.} Mean metrics over three runs. Bold and underlined values denote the best and second-best mean, respectively.}
    \label{tab:celeba_vae_results}
    \resizebox{\columnwidth}{!}{%
    \begin{tabular}{@{}lcc|cccc@{}}
    \toprule
    \textbf{Method} & \textbf{$\lambda_{\mathrm{recon}}$} & \textbf{$\lambda_{\mathrm{kld}}$} & \textbf{rFID} $\downarrow$ & \textbf{SSIM} $\uparrow$ & \textbf{SSNR} $\uparrow$ & \textbf{PSNR} $\uparrow$ \\
    \midrule
    VAE + LS & 1 & 0.00025 & 78.9695 & 0.65763 & 8.2857 & 20.4554 \\
    \midrule
    VAE + MGDA & 1 & 0.00025 & 296.1916 & 0.37072 & 0.1651 & 12.3348 \\
    VAE + MGDA-GN & 1 & 0.00025 & 77.8938 & 0.65705 & 8.2626 & 20.4323 \\
    VAE + MGDA-LN & 1 & 0.00025 & \textbf{46.6473} & \textbf{0.78999} & \textbf{12.1807} & \textbf{24.3504} \\
    VAE + MGDA-LGN & 1 & 0.00025 & \underline{48.0444} & \underline{0.78483} & \underline{11.9713} & \underline{24.1410} \\
    VAE + AMTL-min & 1 & 0.00025 & 109.6532 & 0.55572 & 5.8271 & 17.9968 \\
    VAE + UPGrad & 1 & 0.00025 & 121.9887 & 0.51188 & 4.5761 & 16.7458 \\
    \bottomrule
    \end{tabular}
    }
\end{table}

\begin{figure}[!t]
    \centering
    \includegraphics[width=0.95\columnwidth]{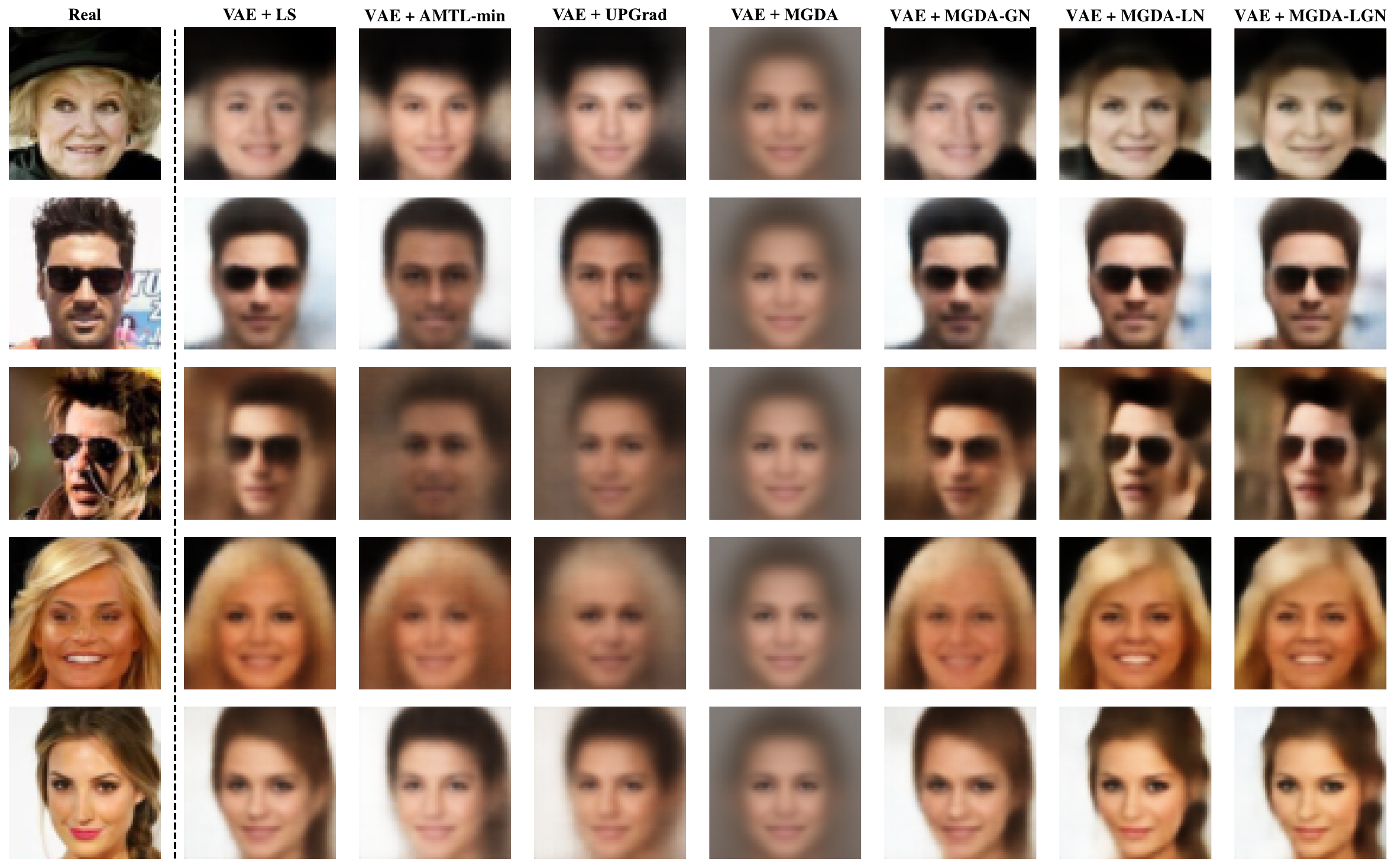}
    \caption{\textbf{Reconstruction on CelebA.} Each row shows a test image (leftmost, ``Real'') and its reconstructions by different methods.}
    \label{fig:celeba_vae_recon}
\end{figure}

\Cref{tab:celeba_vae_results,fig:celeba_vae_recon} show a strong dependence on the aggregation variant. MGDA-LN reduces reconstruction FID (rFID) from 78.97 to 46.65 and increases PSNR from 20.46 to 24.35 under the reported CelebA configuration, with improvements in SSIM and SSNR as well. MGDA-LGN gives the second-best values. In contrast, vanilla MGDA, UPGrad, and AMTL-min underperform LS on all four reconstruction metrics. Loss normalization is therefore consequential in this experiment, but the comparison does not isolate normalization from all other changes in the effective update scale.

\begin{table*}[!t]
    \centering
    \caption{\textbf{VAE evaluation comparison on CelebA-HQ.} We compare reconstruction objectives and gradient aggregation methods using rFID, gFID, and LPIPS (lower is better), and SSIM and PSNR (higher is better). The best value for each metric is shown in bold and the second-best is underlined.}
    \label{tab:celeba_hq_vae_results_csv}
    {\small\setlength{\tabcolsep}{6pt}
    \begin{tabular}{@{}lccc|ccccc@{}}
    \toprule
    \textbf{Method} & \textbf{$\mathcal{L}_{\mathrm{recon}}$} & \textbf{$\lambda_{\mathrm{recon}}$} & \textbf{$\lambda_{\mathrm{kld}}$} & \textbf{rFID} $\downarrow$ & \textbf{gFID} $\downarrow$ & \textbf{SSIM} $\uparrow$ & \textbf{LPIPS} $\downarrow$ & \textbf{PSNR} $\uparrow$ \\
    \midrule
    VAE + LS & BCE & 1 & 0.00457 & 188.9280 & 192.4487 & 0.48962 & 1.2386 & 15.2778 \\
    VAE + AMTL-median & BCE & 1 & 0.00457 & 210.0220 & 218.1875 & 0.48025 & 1.2496 & 14.8144 \\
    VAE + AMTL-min & BCE & 1 & 0.00457 & 207.1659 & 215.0438 & 0.48091 & 1.2486 & 14.8519 \\
    VAE + MGDA-LGN & BCE & 1 & 0.00457 & 143.4755 & 152.3432 & \textbf{0.53859} & \textbf{1.1678} & \textbf{17.7619} \\
    \midrule
    VAE + LS & MSE & 1 & 0.00457 & 149.5810 & 153.5041 & 0.49770 & 1.2285 & 15.9448 \\
    VAE + UPGrad & MSE & 1 & 0.00457 & 214.1389 & 217.7569 & 0.46770 & 1.2677 & 14.2731 \\
    VAE + AMTL-min & MSE & 1 & 0.00457 & 161.7625 & 163.7275 & 0.49020 & 1.2375 & 15.5183 \\
    VAE + AMTL-median & MSE & 1 & 0.00457 & 161.3476 & 163.5512 & 0.49015 & 1.2372 & 15.5197 \\
    VAE + MGDA-LN & MSE & 1 & 0.00457 & 149.0707 & 152.4832 & 0.49755 & 1.2285 & 15.9305 \\
    VAE + MGDA-LGN & MSE & 1 & 0.00457 & \underline{137.2693} & \underline{141.6053} & 0.51024 & 1.2103 & 16.6240 \\
    VAE + MGDA-GN & MSE & 1 & 0.00457 & \textbf{129.4437} & \textbf{134.9142} & \underline{0.51772} & \underline{1.1985} & \underline{17.0551} \\
    \bottomrule
    \end{tabular}
    }
\end{table*}

\begin{figure*}[!t]
    \centering
    \subfloat[CelebA\label{fig:celeba_vae_gradient_similarity}]{\includegraphics[width=0.49\textwidth]{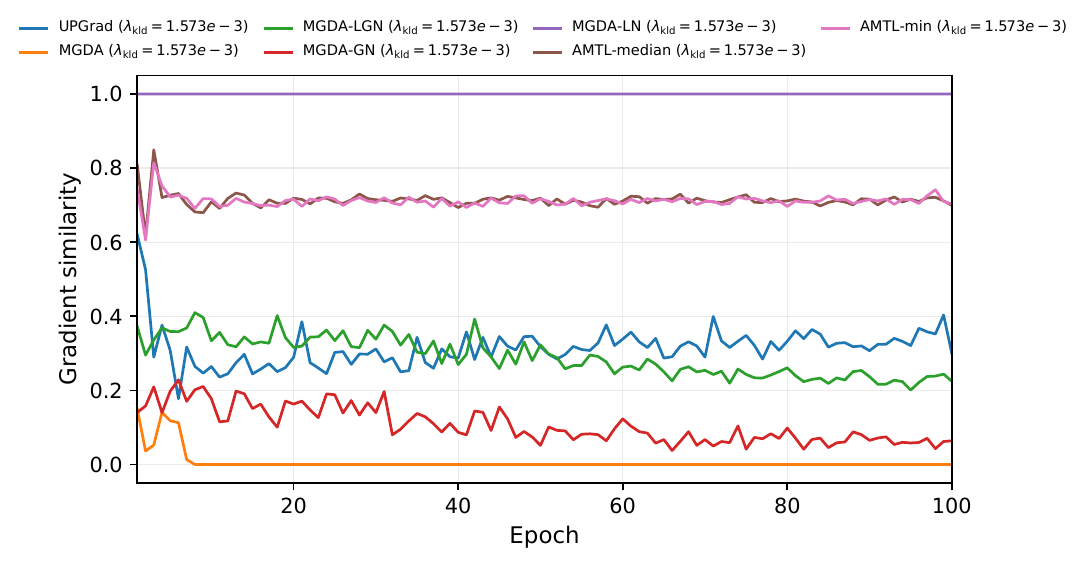}}
    \hfill
    \subfloat[CelebA-HQ\label{fig:celebahq_vae_gradient_similarity}]{\includegraphics[width=0.49\textwidth]{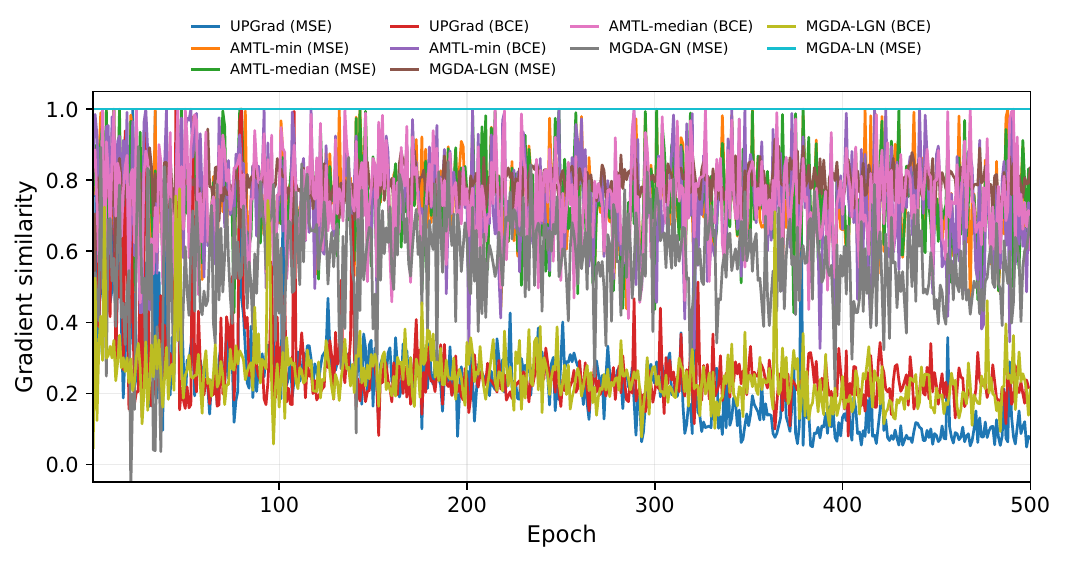}}
    \caption{\textbf{Gradient similarity during VAE training.} Each curve reports the cosine similarity between the MOO aggregated update direction and the LS update direction over epochs. The CelebA panel uses one aggregation configuration per curve, while the CelebA-HQ panel uses one aggregation and reconstruction loss configuration per curve.}
    \label{fig:vae_gradient_similarity}
\end{figure*}

The CelebA-HQ comparison in \cref{tab:celeba_hq_vae_results_csv} uses $\lambda_{\mathrm{kld}}=0.00457$ and evaluates both BCE and MSE reconstruction losses. It further limits a universal recommendation: with MSE, MGDA-GN gives the lowest reconstruction and generation FID (rFID and gFID), whereas with BCE, MGDA-LGN improves on LS in all five metrics. The AMTL variants and UPGrad underperform LS in both configurations. These outcomes support evaluating normalization together with the reconstruction objective. Generation FID follows reconstruction FID closely in this comparison, but reconstruction gains alone do not establish better latent representations.

\Cref{fig:vae_gradient_similarity} shows how far each aggregation method rotates the update away from the LS direction during training. MGDA-LN keeps the LS direction on both datasets (similarity 1.0), so it differs from LS only in the scale of each update. Vanilla MGDA on CelebA drops to zero similarity within the first ten epochs, consistent with its collapsed reconstruction quality. The remaining methods rotate the update substantially: on CelebA, the AMTL variants stay near 0.7, while UPGrad, MGDA-LGN, and MGDA-GN fall to between 0.05 and 0.4. The amount of rotation does not by itself predict reconstruction quality; on CelebA, MGDA-LGN improves on LS despite a low similarity, whereas UPGrad does not.

We also evaluate the vector-quantized VAE (VQ-VAE)~\cite{Oord2017}, whose training combines three objectives: reconstruction, vector quantization (codebook) loss, and commitment loss. Gradient-Guided VQ-VAE (GG-VQ-VAE) adds two auxiliary objectives targeting high-frequency image structure: a reconstruction loss weighted by the normalized Sobel gradient magnitude of the input, and an $\ell_1$ edge-matching loss between the Sobel gradient magnitudes of the input and its reconstruction. This gives three and five objectives, respectively, whose objective-wise gradients are aggregated in the same way as for the VAE. The loss definitions and training protocol are given in Appendix~\ref{S-app:gg_vqvae} in the supplementary material.

\begin{table*}[!t]
    \centering
    \caption{\textbf{VQ-VAE reconstruction evaluation metrics on ImageNet.} Comparison of LS and UPGrad for training VQ-VAE and GG-VQ-VAE models. We report rFID (lower is better), SSIM, SSNR, and PSNR (higher is better). The best result for each metric is shown in bold.}
    \label{tab:imagenet_results}
    {\small\setlength{\tabcolsep}{8pt}
    \begin{tabular}{@{}lccccc|cccc@{}}
    \toprule
    Method & $\lambda_{\mathrm{recon}}$ & $\lambda_{\mathrm{vq}}$ & $\lambda_{\mathrm{cm}}$ & $\lambda_{\mathrm{gg}}$ & $\lambda_{\mathrm{em}}$ & rFID $\downarrow$ & SSIM $\uparrow$ & SSNR $\uparrow$ & PSNR $\uparrow$ \\
    \midrule
    VQ-VAE + LS & 1 & 1 & 0.25 & -- & -- & 14.653 & 0.813 & \textbf{14.145} & 26.499 \\
    VQ-VAE + UPGrad & 1 & 1 & 0.25 & -- & -- & 17.147 & 0.795 & 13.520 & 26.344 \\
    \midrule
    GG-VQ-VAE + LS & 1 & 1 & 0.25 & 1 & 1 & 17.239 & 0.811 & 13.090 & 25.938 \\
    GG-VQ-VAE + UPGrad & 1 & 1 & 0.25 & 1 & 1 & \textbf{13.345} & \textbf{0.832} & 13.837 & \textbf{26.795} \\
    \bottomrule
    \end{tabular}
    }
\end{table*}

\begin{figure*}[!t]
    \centering
    \includegraphics[width=0.75\textwidth]{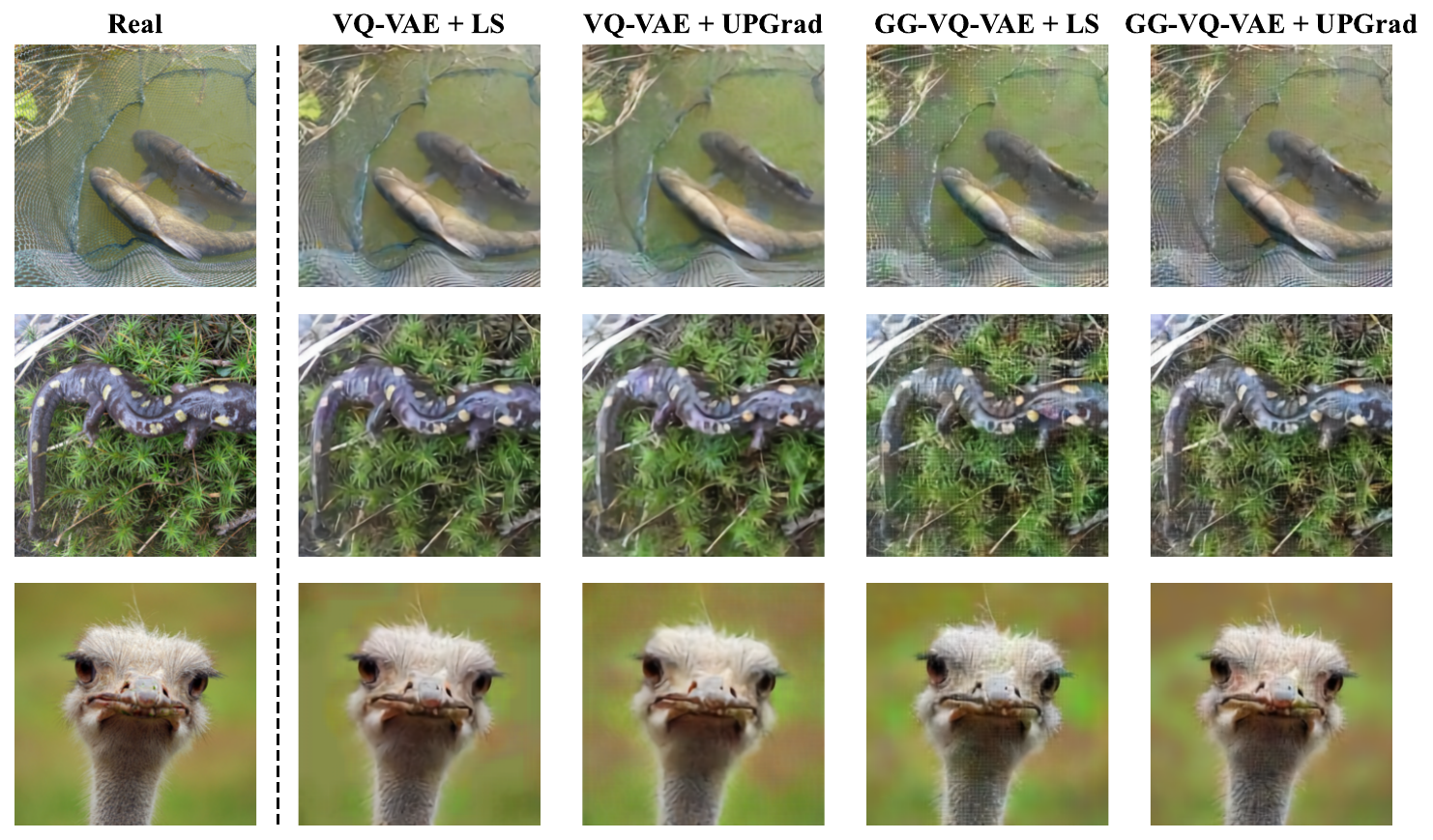}
    \caption{\textbf{VQ-VAE image reconstruction on ImageNet.} Validation data reconstruction comparison between SOO (LS) and MOO (UPGrad) for VQ-VAE and GG-VQ-VAE after training on ImageNet.}
    \label{fig:recon_imagenet}
\end{figure*}

Looking at \cref{tab:imagenet_results,fig:recon_imagenet} for ImageNet, UPGrad improves all four reported reconstruction metrics over LS for GG-VQ-VAE, reducing rFID from 17.239 to 13.345. However, it worsens all four metrics for the base VQ-VAE. Thus, even within one model family and dataset, the benefit of an aggregation method depends on the objective set. Comparisons of LS and aggregation within the same formulation are needed to distinguish this effect from the effect of adding auxiliary losses. The additional CelebA and CelebA-HQ results in Appendix~\ref{S-app:gg_vqvae} also show that rankings do not transfer from the continuous VAE: the MGDA variants that improved VAE reconstruction substantially degrade VQ-VAE and GG-VQ-VAE reconstruction on CelebA-HQ.

\section{Multi-Objective Reinforcement Learning for Molecular Optimization}
\label{sec:MORL}

Multi-objective reinforcement learning (MO-RL) treats rewards as a vector rather than a scalar, making conflicts among reward components explicit instead of absorbing them into a weighted reward sum~\cite{parisi2014policy}. This is relevant when a scalar reward combines several properties of the same output, as in molecular optimization, where properties such as drug-likeness, binding, and synthesizability must be satisfied together. We examine molecular lead optimization, where a single policy modifies a molecule to satisfy several property criteria. We first summarize the MO-RL formulation and then apply it to this task.

\subsection{Multi-Objective Markov Decision Processes}
\label{sec:momdp}
The multi-objective Markov decision process (MOMDP)~\cite{parisi2014policy,Kyriakis2022} is defined with seven components: $\{\mathcal{S},\mathcal{A},\mathcal{R},\mathcal{P},\gamma,\Omega,f_{\Omega}\}$. Here, $\mathcal{S}$ is the state space and $\mathcal{A}$ is the action space. $\mathcal{P}:\mathcal{S}\times\mathcal{A}\rightarrow\mathcal{M}(\mathcal{S})$ denotes the state transition kernel, where $\mathcal{M}(\mathcal{S})$ represents the space of all probability distributions defined over $\mathcal{S}$. $\mathcal{R}\equiv\bm{r}(s,a):\mathcal{S}\times\mathcal{A}\rightarrow\mathbb{R}^M$ is a bounded $M$-dimensional reward function, where $M$ is the number of rewards or objectives. For narrative convenience, we assume deterministic reward signals in the following; the formulation extends to stochastic reward distributions. As in an MDP, $\gamma\in(0,1]$ is the discount factor, with $\gamma<1$ required for an infinite horizon. $\Omega\subset\mathbb{R}^M$ represents the space of preferences. A commonly used linear preference space is the probability simplex $\Omega\doteq\{\bm{w}\in\mathbb{R}_{+}^{M}:\bm{w}^\top\bm{1}=1\}$. The preference function $f_{\Omega}:\mathbb{R}^M\rightarrow\mathbb{R}$ produces a scalar utility; the linear preference function is the weighted sum of reward signals $f_{\Omega}(\bm{r}(s,a))=\bm{w}^\top\bm{r}(s,a)$ with $\bm{w}\in\Omega$.

A stationary policy is denoted by $\pi:\mathcal{S}\rightarrow\mathcal{M}(\mathcal{A})$, and $\Pi$ denotes the set of all eligible stationary policies. The corresponding value function over horizon $T$, which may be infinite when $\gamma<1$, is
\begin{equation}
\bm{v}^{\pi}(s)\doteq\mathbb{E}_{\tau\sim p^{\pi}(\tau)}\Big[\sum_{t=1}^{T}\gamma^{t-1}\bm{r}(s_t,a_t)\,\Big|\,s_1=s\Big],
\end{equation}
where $p^{\pi}(\tau)$ is the trajectory distribution induced by $\pi$ and $\mathcal{P}$. Different from single-objective reinforcement learning (SO-RL), the value function is $M$-dimensional: $\bm{v}^{\pi}(s)\in\mathbb{R}^M$. For a discrete state space, $\bm{v}^{\pi}$ has a size of $M\times|\mathcal{S}|$. Defining the discounted state-action occupation measure as $d^{\pi}_{s}(s',a)\doteq\sum_{t=1}^{T}\gamma^{t-1}p^{\pi}(s_t=s',a_t=a\mid s_1=s)$, the value function can be written as $\bm{v}^{\pi}(s)=\sum_{s',a}d^{\pi}_{s}(s',a)\,\bm{r}(s',a)$.

Policies are compared through their expected value vectors $\bm{V}(\pi)\doteq\mathbb{E}_{s\sim\rho_0}[\bm{v}^{\pi}(s)]$ under an initial-state distribution $\rho_0$. A policy $\pi'$ Pareto dominates $\pi$, written $\bm{V}(\pi')\succ_P\bm{V}(\pi)$, if $V_m(\pi')\geq V_m(\pi)$ for every objective $m$, with strict inequality for at least one. The Pareto policy set contains the policies that are not Pareto dominated by any other eligible policy:
\begin{equation}
\Pi_P^*\doteq\{\pi\in\Pi:\nexists\,\pi'\in\Pi\ \text{such that}\ \bm{V}(\pi')\succ_P\bm{V}(\pi)\}.
\end{equation}
For a parameterized policy $\pi_{\theta}$ with differentiable $\bm{V}(\theta)\doteq\bm{V}(\pi_{\theta})$, $\pi_{\theta}$ is Pareto stationary if
\begin{equation}
\sum_{m=1}^M w_m\nabla_{\theta}V_m(\theta)=\bm{0}\quad\text{for some }\bm{w}\in\Omega.
\end{equation}
As in \cref{sec_moo}, this first-order condition is necessary for local Pareto optimality of an unconstrained differentiable policy but does not guarantee it. The vector $\bm{w}$ in this condition certifies stationarity; it is not a preference chosen in advance. Maximizing $\bm{w}^\top\bm{V}(\theta)$ jointly over $\theta$ and $\bm{w}\in\Omega$ would change the decision problem, because an optimal $\bm{w}$ can always be placed at a vertex of the simplex, which reduces the task to maximizing a single objective. When the preference vector is prespecified and fixed, the scalarized task is
\begin{equation}
\max_{\theta}\ f(\theta)=\bm{w}^{\top}\bm{V}(\theta).
\end{equation}
The MO-RL approach evaluated here does not fix such a scalar reward. It estimates the objective-wise policy gradients $\nabla_{\theta}V_m(\theta)$ from objective-specific advantages and aggregates them with a gradient aggregation method from \cref{sec_moo} before each actor update. We next apply this formulation to molecular lead optimization.

\begin{table*}[!t]
    \centering
    \caption{Percentage of molecules satisfying each property from the 1K lead molecules and optimized molecules by different methods. MO-RL results are averaged over 5 independent runs; all other methods over 3 runs (see Appendix~\ref{S-app:drug_design_protocol} in the supplementary material for the protocol).}
    \label{table:lead_opt}
    \begin{tabular}{@{}l|c|cc|ccccc@{}}
        \toprule
        Property & Leads & MO-RL & SO-RL & REINVENT4 & DeepFMPOv3D & DeepFMPO & MolDQN & MIMOSA \\
        \midrule
        Validity & -- & \textbf{100\%} & \textbf{100\%} & \textbf{100\%} & 74.7\% & 62.3\% & \textbf{100\%} & 51.0\% \\
        Novelty & -- & 98.1\% & 88.2\% & \textbf{100\%} & 75.0\% & 68.8\% & \textbf{100\%} & \textbf{100\%} \\
        Uniqueness & -- & 99.0\% & 90.9\% & 99.8\% & 75.0\% & 68.7\% & \textbf{100\%} & 99.8\% \\
        logP ($< 3$) & 73.9\% & 94.2\% & 71.8\% & 96.3\% & 53.6\% & 51.9\% & 92.6\% & \textbf{99.6\%} \\
        SAS ($< 2.5$) & 38.4\% & 72.6\% & \textbf{77.9\%} & 57.6\% & 12.0\% & 16.8\% & 0.2\% & 16.1\% \\
        \makecell[l]{BAS (LPA1)\\($< -7.5$)}& 11.3\% & \textbf{72.9\%} & 47.1\% & 56.1\% & 6.6\% & 7.8\% & 3.7\% & 28.7\% \\
        \midrule
        All Properties & 0.0\% & \textbf{54.1\%} & 23.5\% & 35.9\% & 0.3\% & 0.2\% & 0.0\% & 1.0\% \\
        \bottomrule
    \end{tabular}
\end{table*}

\subsection{Molecular Lead Optimization}
\label{sec:lead_opt}
We use lead optimization as an example to show the effectiveness of MO-RL approaches. Drug discovery and development is a costly, time-consuming, and high-risk process \cite{Sun_2022}. Typically, the process involves high-throughput screening to identify a few lead molecules, lead optimization to improve lead molecules' physicochemical properties, preclinical trials, and clinical trials. At each such step, the majority of drug candidates are filtered out for not meeting certain physicochemical criteria, such as solubility, toxicity, half-life, and efficacy. To date, only about 2,000 molecules have been approved for use as medicines \cite{Wishart2018}. The rise of deep learning has dramatically influenced computational drug design in the past decade. Based on sequence embedding \cite{GmezBombarelli2018} and graph representation \cite{Duvenaud2015}, autoencoders \cite{Jin2018}, Transformers \cite{castro2022transformer}, and diffusion models \cite{schneuing_2024}, have been developed for \textit{molecular generation} which aims at learning the distribution of the molecular data used in training and then generating novel compounds that can then be used for subsequent screening. Another active stream of research is \textit{molecular optimization}, which aims to (slightly) modify given molecular structures to improve multiple properties. Specifically in the jargon of drug discovery, it is called \textit{lead optimization}. RL has been considered to tackle this task, such as MolDQN \cite{MolDQN}, DeepFMPO \cite{deepFMPO}, MIMOSA \cite{fu_2021}, DeepFMPOv3D \cite{deepFMPOv3D}, and REINVENT4 \cite{loeffler_2024}, among others. 
However, these RL approaches use a single reward signal and do not account for potential conflicts among multiple properties (objectives), leading to unsatisfactory results. We hypothesize that MO-RL can perform much better. In our preliminary research, we developed an advantage-based actor-critic model using a Transformer (pretrained using molecules from the MolGen task of Therapeutics Data Commons (TDC) \cite{TDC}) for optimizing a given set of 1,000 molecules that do not satisfy all property requirements including logP $<3$, synthetic accessibility score (SAS) $<2.5$, and binding affinity score (BAS) to a protein target LPA1 \cite{lin_2021} $<-7.5$. The three-objective RL was optimized using AMTL. Among the 1,000 molecules optimized by it, $54.1\%$ of them are in the sweet spot (satisfying all three requirements). Comparison results with SOTA baselines, including REINVENT4 and other SO-RL approaches, are shown in \cref{table:lead_opt}, demonstrating that MO-RL with explicit conflict resolution substantially outperforms methods that rely on single scalarized rewards. 

\section{Discussion and Conclusion}\label{sec:discussion}
\subsection{When Objective-Wise Optimization Is Useful}
The experiments support a conditional use of MOO in single-task learning. Meaningful objective components make trade-offs explicit, but decomposition alone does not establish a benefit over scalarization. In classification, much of the improvement over ERM is already obtained by class-balanced CWRM+LS. In generative modeling, changing the normalization or objective set can reverse the ranking of aggregation methods. In segmentation and molecular optimization, the preferred solution depends on which combinations of evaluation criteria matter for the application.

A practical starting point is therefore a scalarized baseline with a defined validation criterion and a documented budget for tuning its weights and learning rate. Objective-wise losses, gradient norms, and the inner products in \Cref{sec:local_diagnostic} can then identify persistent imbalance or adverse local updates. When these patterns accompany unsatisfactory validation trade-offs, a limited comparison of aggregation methods is warranted. A negative cosine similarity by itself is not a reason to replace LS because the trade-off may be intended, the estimated gradients may be noisy, or aggregation may reduce training interference without improving generalization. If the baseline already meets the application requirements, the experiments provide no basis for assuming that additional MOO machinery will be beneficial.

\subsection{Selecting and Evaluating an Aggregation Method}
Our results suggest starting configurations for comparable settings, rather than a universal ranking. For the two-objective CelebA VAE, MGDA-LN and MGDA-LGN are useful candidates; the CelebA-HQ results show that the reconstruction loss and dataset can change which normalization works best. For CIFAR100-LT, UPGrad and AMTL-median/RMSE are stronger starting points than the tested unnormalized MGDA or AMTL-min configurations. The ImageNet comparison cautions against transferring even a successful choice across objective sets: UPGrad helps GG-VQ-VAE and hurts base VQ-VAE. These observations are candidates for validation on a new problem, not rules inferred from objective count alone.

Objective scaling and optimizer settings are part of method selection. Loss magnitudes and gradient norms convey different information; for instance, a large loss value need not imply a large gradient, and a small gradient norm need not indicate a difficult objective. The MGDA normalization variants change the geometry supplied to the aggregation subproblem, while AMTL scaling changes update magnitudes. Apparent benefits can therefore reflect both direction and effective step size. Momentum and adaptive preconditioning introduce further changes after aggregation. Comparisons that hold the underlying model and data protocol fixed, tune learning rates comparably, and examine actual parameter updates are needed to distinguish these effects.

Model selection also requires a declared target. For a single deployed model, this may be a primary validation metric subject to thresholds on other criteria, as illustrated by joint property satisfaction in molecular optimization. When several operating points are useful, training with a small set of specified preferences and reporting the empirically non-dominated validation outcomes offers a practical way to compare trade-offs. Such a set is not a certified Pareto front. The present experiments principally compare individual trained models and do not evaluate systematic front coverage.

\subsection{Computational Cost and Scalability}
\begin{table}[!t]
\centering
\caption{\textbf{Training runtime for selected configurations.} Times in hours are averaged over three seeds.}
\label{tab:runtime}
{\footnotesize\setlength{\tabcolsep}{4pt}
\begin{tabular}{llrrr}
\toprule
Setting & Method & $M$ & Time (h) & Rel. \\
\midrule
\makecell[l]{CelebA-HQ\\VAE (MSE)} & LS & 2 & 3.14 & $1.00\times$ \\
& AMTL-median & 2 & 3.93 & $1.25\times$ \\
& AMTL-min & 2 & 3.96 & $1.26\times$ \\
& UPGrad & 2 & 4.00 & $1.27\times$ \\
& MGDA-LGN & 2 & 4.01 & $1.28\times$ \\
\midrule
\makecell[l]{CelebA-HQ\\VQ-VAE (MSE)} & LS & 3 & 6.30 & $1.00\times$ \\
& AMTL-median & 3 & 10.25 & $1.63\times$ \\
& MGDA-LN & 3 & 10.24 & $1.63\times$ \\
& MGDA-GN & 3 & 10.24 & $1.63\times$ \\
& MGDA-LGN & 3 & 10.31 & $1.64\times$ \\
\midrule
\makecell[l]{CIFAR100-LT\\(IF=100)} & ERM & 1 & 0.42 & $1.0\times$ \\
& CWRM+LS & 100 & 2.11 & $5.0\times$ \\
& CWRM+AMTL-median & 100 & 2.15 & $5.1\times$ \\
& CWRM+MGDA-LN & 100 & 2.28 & $5.4\times$ \\
& CWRM+UPGrad & 100 & 2.32 & $5.5\times$ \\
\bottomrule
\end{tabular}
}
\end{table}

\Cref{tab:runtime} reports available runtime measurements for selected configurations; their provenance is described in Appendix~\ref{S-app:runtime} in the supplementary material. Relative to their respective LS baselines, the measured CelebA-HQ VAE runs take approximately $1.25$--$1.28\times$ as long, and the measured VQ-VAE runs take $1.63$--$1.64\times$ as long. For CIFAR100-LT at IF=100, CWRM+LS takes approximately $5.0\times$ the ERM runtime, while the listed aggregation methods take $5.1$--$5.5\times$. Most of the measured classification overhead therefore comes from the CWRM implementation itself, and gradient aggregation adds only about $2$--$10\%$ on top of CWRM+LS. The fivefold cost is not inherent to class-balanced scalarization, which can in principle be computed as a single scalar loss.

These measurements describe the evaluated implementations, not asymptotic laws or equal-compute comparisons. They do not include peak memory measurements or total hyperparameter search costs. Materializing all objective gradients requires $O(MN)$ storage for $M$ objectives and $N$ parameters. Methods based on the $M\times M$ Gram matrix $\bm{J}\bm{J}^{\top}$ can separate the coefficient computation from the final gradient combination~\cite{quinton2024jacobian}, but constructing this matrix and recovering the update still incurs computation. Reduced gradient storage does not imply single-objective runtime. The trade-off between memory, recomputation, and solver cost becomes especially relevant as the number of objectives grows.

\subsection{Limitations and Further Evaluation}
The study covers selected datasets, decompositions, and aggregation methods, with different method coverage across applications. The reported means do not quantify uncertainty, and small differences, particularly in classification, require run-level variability to assess their reliability. Weight and normalization choices are not exhaustively searched, and the experiments do not establish superiority over optimally tuned scalarization or under a matched total compute budget. The segmentation evaluation uses spatial and morphological measures that do not independently establish boundary accuracy. Likewise, the generative model reconstruction results and molecular property scores support conclusions about those metrics, rather than about all aspects of generation or downstream biological utility.

The local analysis applies to the supplied objective gradients. Stochastic estimates, changing gradient scales, and optimizer transformations limit its direct interpretation as a statement about the full-data objectives or generalization. The experiments also do not address all forms of data heterogeneity, nonstationary learning, or environmental conflict. Further evaluation can test whether gradient diagnostics predict benefits on held-out settings, isolate normalization from step-size effects, and compare solutions under matched tuning and training budgets. Memory gradient computation with efficiency and preference conditioned training are additional directions for scaling the approach and assessing trade-off coverage.

\subsection{Conclusion}
This study examined multi-objective formulations of single-task learning across classification, cell segmentation, generative modeling, and reinforcement learning. Objective-wise optimization improved selected performance measures and joint property satisfaction relative to the evaluated scalarized baselines, while the negative results demonstrated substantial dependence on objective decomposition, normalization, and aggregation method. The evidence supports MOO as a useful option when a learning problem contains meaningful competing criteria and the resulting trade-offs matter for validation or deployment. It also shows why method selection must account for scalarized baselines, optimizer settings, and computational cost. In particular, the same aggregation method can improve an augmented objective set and degrade the corresponding base model, making objective construction and aggregation closely related to design choices. Establishing predictive selection criteria and efficient implementations is a natural next step toward more reliable use of these methods.

\section*{Acknowledgment}
This work was supported in part by the Natural Sciences and Engineering Research Council of Canada (NSERC) through Discovery Grants awarded to B. Ombuki-Berman, S. Rahnamayan, and Y. Li. Y. Li was also supported by the Canada Research Chairs Program and the Canada Foundation for Innovation (CFI) John R. Evans Leaders Fund.

\onecolumn
\section*{Supplementary Material}
\setcounter{table}{0}
\setcounter{figure}{0}
\setcounter{equation}{0}
\renewcommand{\theHtable}{supp.\arabic{table}}
\renewcommand{\theHfigure}{supp.\arabic{figure}}
\renewcommand{\theHequation}{supp.\arabic{equation}}
\renewcommand{\thetable}{S\arabic{table}}
\renewcommand{\thefigure}{S\arabic{figure}}
\renewcommand{\theequation}{S\arabic{equation}}

This document supplements the main paper. Sections, tables, figures, and equations without the prefix ``S'' or ``Appendix'' refer to the main paper.

\appendices

\section*{Contents}
\startcontents[appendix]
\printcontents[appendix]{}{1}{\setcounter{tocdepth}{2}}

\newpage
\section{Gradient Aggregation Methods and Diagnostics}
\label{S-app:gradient_methods}

\subsection{Gradient Aggregation Variants}
\label{S-app:agg_variants}
\paragraph{MGDA and its normalization variants}
Let $\{\bm{g}_m\}_{m=1}^M$ denote per-objective gradients (flattened into vectors). MGDA computes a convex combination by solving for weights $\bm{\alpha}^*\in\Delta^{M-1}$ (the probability simplex) that minimize the norm of the aggregated update:
\begin{equation}
\bm{\alpha}^* = \arg \min_{\bm{\alpha}\in\Delta^{M-1}}\left\|\sum_{m=1}^M \alpha_m \tilde{\bm{g}}_m\right\|_2^2,
\end{equation}
where $\tilde{\bm{g}}_m$ is a normalized version of $\bm{g}_m$ (normalization is applied {before} solving the simplex problem).
The aggregated gradient is then $G_{\mathcal{A}_{\text{MGDA}}} = \sum_{m=1}^M \alpha_m^* \tilde{\bm{g}}_m$.
There are three suggested normalization techniques by \cite{lin2023libmtl} as follows:
\begin{itemize}
    \item MGDA-GN: $\tilde{\bm{g}}_m=\bm{g}_m/\|\bm{g}_m\|_2$, normalize by gradient norm.
    \item MGDA-LN: $\tilde{\bm{g}}_m=\bm{g}_m/L_m$, normalize by loss magnitude.
    \item MGDA-LGN: $\tilde{\bm{g}}_m=\bm{g}_m/(L_m\|\bm{g}_m\|_2)$, normalize by both loss magnitude and gradient norm.
\end{itemize}
Here, $L_m$ denotes the loss magnitude used for normalization. These variants rescale objective gradients before the MGDA subproblem and can change both the direction and magnitude of the resulting update. Their performance is assessed empirically; positive rescaling alone does not ensure an improvement over LS.

\paragraph{AMTL scaling modes}
AMTL \cite{senushkin2023independent} aligns objective-wise gradients by replacing the singular values of the gradient matrix with a common scale.
Let $\bm{J}\in\mathbb{R}^{M\times N}$ stack per-objective gradients as rows, so that $\bm{J}^\top=[\bm{g}_1,\ldots,\bm{g}_M]$.
Given a compact SVD on the rank-$r$ subspace, $\bm{J}^\top=\bm{U}_r\bm{\Sigma}_r\bm{V}_r^\top$, AMTL constructs the following:
\begin{equation}
\widetilde{\bm{J}}^\top = \bm{U}_r(s\bm{I}_r)\bm{V}_r^\top,
\end{equation}
where the columns of $\widetilde{\bm{J}}^\top$ are the aligned gradients $\widetilde{\bm{g}}_m$.
The aggregated gradient is then computed as $G_{\mathcal{A}_{\text{AMTL}}}=\sum_{m=1}^{M}\widetilde{\bm{g}}_m$.
We consider the following scale choices for $s$:
\begin{itemize}
    \item AMTL-min: $s = \min_i^+(\sigma_i)$, the smallest positive singular value.
    \item AMTL-median: $s = \mathrm{median}_i^+(\sigma_i)$, the median positive singular value.
    \item AMTL-RMSE: $s = \sqrt{\frac{1}{r}\sum_{i=1}^{r}\sigma_i^2}$, the root-mean-square of the non-zero singular values, where $r=\mathrm{rank}(\bm{J})$.
\end{itemize}
The minimum (min) is sensitive to the smallest retained singular value, the median uses a central value, and RMSE reflects the overall magnitudes of the singular values. These scaling choices alter the effective step size and are evaluated separately in the experiments.

\subsection{Gradient Diagnostic Metrics}
\label{S-app:gradient_diagnostics}
The following diagnostics summarize objective-wise gradient geometry during training.
For a given update step, let $\bm{g}_m\in\mathbb{R}^{N}$ denote the gradient of objective $m$, and let $\bm{J}\in\mathbb{R}^{M\times N}$ stack the objective-wise gradients as rows.
The LS baseline produces
\begin{equation}
\bm{G}_{\text{LS}}=\sum_{m=1}^{M}\bm{g}_m,
\end{equation}
whereas an MOO aggregation method produces
\begin{equation}
\bm{G}_{\mathcal{A}}=\mathcal{A}(\bm{J}),
\end{equation}
with $\bm{J}$ stacking the objective-wise gradients as rows.

\paragraph{Gradient similarity}
The cosine similarity metric between the MOO aggregated direction and the SOO LS direction is calculated as:
\begin{equation}
\mathrm{GradSim}
    =
\frac{\bm{G}_{\mathcal{A}}^\top \bm{G}_{\text{LS}}}
{\|\bm{G}_{\mathcal{A}}\|_2\,\|\bm{G}_{\text{LS}}\|_2}.
\end{equation}
\begin{itemize}
    \item $\mathrm{GradSim}\approx 1$: the MOO aggregated update closely aligns with the LS direction, meaning the aggregation method introduces little rotation relative to the SOO baseline.
    \item $\mathrm{GradSim}\approx 0$: the two directions are nearly orthogonal, indicating that the aggregation method is substantially redirecting the update away from the LS direction.
\end{itemize}

\paragraph{Gradient inner product}
To summarize pairwise agreement, define $\mathcal{P}=\{(i,j):1\leq i<j\leq M\}$.
The metric $\mathrm{GradInnerProd}$ averages all off-diagonal gradient inner products,
\begin{equation}
\mathrm{GradInnerProd}
=
\frac{1}{|\mathcal{P}|}\sum_{(i,j)\in\mathcal{P}}\bm{g}_i^\top\bm{g}_j .
\end{equation}
\begin{itemize}
    \item $\mathrm{GradInnerProd}>0$: positive pairwise inner products dominate the average. This does not imply that every pair agrees or that the aggregated update decreases every loss. 
    \item $\mathrm{GradInnerProd}<0$: objectives are in average conflict, meaning progress on one objective tends to oppose progress on others.
    \item $\mathrm{GradInnerProd}\approx 0$: the average pairwise interaction is small. This may reflect near-orthogonal gradients, small gradient norms, or cancellation between large positive and negative inner products, so it does not by itself imply pairwise orthogonality.
\end{itemize}
\paragraph{Gradient conflict energy}
Finally, $\mathrm{GradConflictEnergy}$ records only the negative part of these pairwise interactions:
\begin{equation}
\mathrm{GradConflictEnergy}
=
\frac{1}{|\mathcal{P}|}\sum_{(i,j)\in\mathcal{P}}\max\left(0,-\bm{g}_i^\top\bm{g}_j\right).
\end{equation}
\begin{itemize}
    \item $\mathrm{GradConflictEnergy}=0$: no pair of objective gradients has a negative inner product, so all objectives are mutually non-conflicting at this step.
    \item $\mathrm{GradConflictEnergy}>0$: at least one pair of objectives conflicts. The value depends on gradient magnitude as well as the number and strength of negative interactions. A larger value indicates stronger or more widespread destructive pairwise interference.
\end{itemize}

\subsection{Runtime Measurements}
\label{S-app:runtime}
\Cref{tab:runtime} in \cref{sec:discussion} of the main paper reports average training runtime over three seeds for the listed configurations. The two CelebA-HQ timing experiments use MSE reconstruction, and each has a separate LS reference. The classification measurements use ERM as their reference. The available timing summary provides only relative runtimes for the listed CWRM variants. These timings are specific to the configuration and do not reflect the full experimental or hyperparameter search budget. The MSE VQ-VAE timing configuration is reported separately from the reconstruction-loss protocol described in Appendix~\ref{S-app:gg_vqvae}.

\section{Long-Tailed Classification Protocol}
\label{S-app:classification_details}
\label{S-app:longtail}

\paragraph{Model architecture} We employ ResNet-32 \cite{he2016deep} as the backbone architecture for all experiments on CIFAR10-LT and CIFAR100-LT.

\paragraph{Training protocol} Following standard practices in long-tailed recognition literature \cite{cui2019class}, we train for 200 epochs with a batch size of 256. Optimization is performed using stochastic gradient descent (SGD) with momentum 0.9 and weight decay $5\times10^{-4}$. The initial learning rate is set to 0.1 and decayed by a factor of 0.1 at epochs 160 and 180 using a multi-step learning rate scheduler.

\paragraph{Class-balanced sampling} For CWRM methods, we adopt class-balanced sampling to reduce the influence of unequal class frequencies when computing class-wise gradients. This contrasts with standard ERM training, which samples uniformly from the imbalanced training set.

\paragraph{Dataset construction} Long-tailed versions of CIFAR10 and CIFAR100 are constructed by reducing the number of training samples per class exponentially according to the imbalance factor $\rho$. Specifically, if class $k$ has $n_k$ samples, we set $n_k = n_{\max} \cdot \rho^{-k/(K-1)}$ for $k\in\{0,\ldots,K-1\}$, where $n_{\max}$ is the number of samples in the most frequent class and $K$ is the total number of classes.

\paragraph{Computational resources} All experiments are conducted on a single NVIDIA A100-SXM4-80GB GPU. Each experiment is repeated over 3 independent runs with different random seeds, and the results in \cref{tab:cifar_lt} of the main paper report the mean performance metrics.

\section{Cell Segmentation Protocol and Datasets}
\label{S-sec_exp_moseg}
\paragraph{Model architecture} For this study, the core architecture of BIDCell remained unchanged, but we performed significant refactoring of the code base to allow for interchangeability of use of different backpropagation solvers and analytical pipelines \cite{fu_bidcell_2024}. BIDCell's training algorithm relies on 6 loss terms. These are summed into a primary loss function using linear scalarization and handled within one large training loop. The individual blocks within the training loop were refactored into their own functions within the BIDCell pipeline for more versatility.

\paragraph{Training protocol}

The experiments were performed with $5$ repeats each. Preference weights for the objectives are set equally. The results from the different repeats have been averaged. Each training was carried out in 7920 steps. Different learning rates (LRs) were explored, and finally, we used $\text{LR}=10^{-6}$ for the reported experiments of all methods here.

We evaluated the MOO algorithms on three breast cancer spatial transcriptomics datasets spanning different acquisition platforms: 1) Xenium (10x Genomics) \cite{xenium_janesick_2023}, 2) MERFISH (Vizgen MERSCOPE) \cite{merfish_chen_2015}, and 3) an in-house NanoString CosMx SMI data. These three datasets allow us to assess the algorithms under distinct spatial transcriptomics platforms and acquisition conditions. Xenium and MERFISH serve as public reference datasets for reproducible benchmarking, while the CosMx cohort provides an independent institutional dataset for evaluating segmentation under a different acquisition platform.  \cref{S-tab:data_summary} summarizes the key information of the three datasets.

\begin{table}[!htbp]
\centering
\small
\caption{Meta information of the datasets used in the multi-objective cell segmentation (MOSeg) experiments.}
\label{S-tab:data_summary}
\resizebox{\linewidth}{!}{%
\begin{tabular}{lrrrrrrrr}
\toprule
Dataset & \makecell{Total\\Transcripts} & \makecell{Area\\(mm$^2$)} & Genes & \makecell{Native px\\($\mu$m)} & \makecell{Target px\\($\mu$m)} & \makecell{Image size\\(px)} & \makecell{Total\\patches} & \makecell{Testing\\patches} \\
\midrule
Xenium  & 43,585,817 & 41.12  & 313 & 0.2125 & 1.0 & $5{,}478\times7{,}526$   & 18,170  & 3,634 \\
MERFISH & 59,731,758 & 19.47  & 815 & 0.108  & 1.0 & $4{,}640\times4{,}242$   &  8,730  & 1,746 \\
CosMx    & 13,482,021 & 124.53 & 960 & 0.500  & 0.5 & $18{,}240\times27{,}360$ & 218,694 & 43,739 \\
\bottomrule
\end{tabular}}
\end{table}

\paragraph{Self-supervised segmentation objectives}
The six objectives in \cref{eq:sscs} of the main paper are adapted from BIDCell~\cite{fu_bidcell_2024} which are as follows:
1) nuclei encapsulation ($\mathcal{L}_{\mathrm{ne}}$), 2) cell calling ($\mathcal{L}_{\mathrm{cc}}$), 3) over-segmentation ($\mathcal{L}_{\mathrm{os}}$), 4) multiple assignment ($\mathcal{L}_{\mathrm{mu}}$, which replaces BIDCell's overlap loss), and 5) positive and 6) negative marker losses ($\mathcal{L}_{\mathrm{pos}}$, $\mathcal{L}_{\mathrm{neg}}$). LS sums them into a single loss form, whereas the MOO methods aggregate their objective-wise gradients. Loss normalization equalizes loss values, not necessarily gradient norms, so a single objective's gradient can still dominate the common direction update. Each objective is defined as follows:

\textit{Nuclei encapsulation loss ($\mathcal{L}_{\mathrm{ne}}$):}
\begin{equation}
    \mathcal{L}_{\mathrm{ne}}(x_{\mathrm{nuc}},\hat{y})=-x_{\mathrm{nuc}}\log(\hat{y})-(1-x_{\mathrm{nuc}})\log(1-\hat{y}),
\end{equation} 
where $x_{\mathrm{nuc}}$ is the binary nucleus segmentation mask, and $\hat{y}$ is the predicted segmentation for all cells of the corresponding training patch. This binary cross-entropy loss encourages agreement with the nucleus mask and is formulated as a penalty rather than a hard containment constraint.

\textit{Cell calling loss ($\mathcal{L}_{\mathrm{cc}}$):}
\begin{equation}
    \mathcal{L}_{\mathrm{cc}}(e,\hat{y})=\frac{1}{C}\sum_{c=1}^C \big\{ -e_c\log(\hat{y}_c)-(1-e_c)\log(1-\hat{y}_c)\big\}, 
\end{equation}
where $e_c$ is the expansion mask and $\hat{y}_c$ is the predicted segmentation of cell $c$ of $C$ cells in an input patch. This loss encourages transcripts around each nucleus to be captured within that nucleus's cell mask. Around each nucleus, an expansion region is generated that likely falls inside the actual cell. The model is penalized if the mask fails to cover those areas in the vicinity of the nucleus. This loss also allows the model to generate larger cell sizes when appropriate. 

\textit{Over-segmentation loss ($\mathcal{L}_{\mathrm{os}}$):} 
\begin{align}
\mathcal{L}_{\mathrm{os}}(x_{\mathrm{nuc}},\hat{q})&= \max\big(0, \displaystyle\frac{1}{C}\displaystyle\sum_{c=1}^C(p_{\mathrm{cyt},c}-p_{\mathrm{nuc},c}) \big),
\end{align}
\begin{align}
p_{\mathrm{nuc},c}&=\sum_{i=1}^I\sum_{j=1}^J\sigma\left(\hat{q}_{ijc}x_{\mathrm{nuc},ij}-0.5\right),
\end{align}
\begin{align}
p_{\mathrm{cyt},c}&=\sum_{i=1}^I\sum_{j=1}^J\sigma\big(\hat{q}_{ijc}(1-x_{\mathrm{nuc},ij})-0.5\big),
\end{align}
where $I,J$ are lateral dimensions of the input, for cell $c$ at pixel $(i,j)$ $\hat{q}_{ijc}$ is the predicted foreground probability for cell $c$, $x_{\mathrm{nuc},ij} \in \{0,1\}$ is the binary nucleus mask, and $\sigma$ is the sigmoid function. 
This displayed loss penalizes a positive average excess of soft non-nuclear foreground mass over nuclear foreground mass. Its effect on fragmentation or cell boundaries must be assessed through the resulting segmentations.

\textit{Multiple assignment loss ($\mathcal{L}_{\mathrm{mu}}$):}
\begin{equation}
    \mathcal{L}_{\mathrm{mu}}(E,\hat{q})=\frac{1}{C}\sum_{i=1}^I\sum_{j=1}^J \underbrace{R_{ij}}_{\substack{\text{excess}\\\text{assignments}}}\times\underbrace{E_{ij}}_{\substack{\text{transcript}\\\text{count}}}, 
\end{equation}
where $R_{ij}=\max(A_{ij}-1,0)$ is the penalty term for extra assigned cells above 1, $E_{ij}$ is the transcript count at pixel $(i,j),$ $C$ is the total number of cells, and 
$
    A_{ij}=\sum_{c=1}^C\hat{p}_{ijc}
$
is the sum of all assigned probabilities for the pixel $(i,j)$ across all candidate cells $c$ out of $C$ cells for the current training patch, and
\begin{equation}
    \hat{p}_{ijc}=\frac{\sigma\big(\alpha(\hat{q}_{ijc}-0.5)\big)-\sigma(-0.5\alpha)}{\sigma(0.5\alpha)-\sigma(-0.5\alpha)}
\end{equation}
is the sharpened sigmoid transformation of the predicted cell probability $\hat{q}_{ijc}$, $\alpha$ is the sharpness hyperparameter (set to $10.0$ by default). This transformation function sharpens the input probability $\hat{q}_{ijc}$ and then normalizes it to $[0,1]$. In simple terms, $\mathcal{L}_{\mathrm{mu}}$ checks how many extra cells are assigned to a particular pixel, and penalizes the model if multiple cells are assigned to a single pixel that is rich in transcripts. Note that we replaced BIDCell's overlap loss with this multiple assignment loss, as it did not converge satisfactorily in the evaluated configuration.

\textit{Positive and negative marker losses ($\mathcal{L}_{\mathrm{pos}}, \mathcal{L}_{\mathrm{neg}}$):} 
\begin{equation}
    \mathcal{L}_{\mathrm{pos}}(m_{\mathrm{pos}},\hat{y})=\frac{1}{C}\sum_{c=1}^C \big\{-m_{\mathrm{pos},c}\log(\hat{y}_c)-(1-m_{\mathrm{pos},c})\log(1-\hat{y}_c)\big\},
\end{equation}
\begin{align}
    \mathcal{L}_{\mathrm{neg}}(m_{\mathrm{neg}},\hat{y})=\frac{1}{C}\sum_{c=1}^C\sigma(\hat{q}_cm_{\mathrm{neg},c}-0.5),
\end{align}
where $m_{\mathrm{pos},c}\in\{0,1\}^{I \times J}$ and $m_{\mathrm{neg},c}\in\{0,1\}^{I \times J}$ are sparse maps of encoded positive and negative markers of the cell type for cell $c$. Positive loss uses prior cell-type-specific information (marker genes) to influence the morphology of predicted cells. If prior data indicate that certain genes are expressed in particular cell types, for example, in the cells that tend to be elongated, the model is encouraged to elongate the corresponding cells. The model is rewarded when a marker gene's transcripts lie within the predicted mask of a cell of the corresponding type, and penalized if they lie outside it. Similarly, negative loss penalizes the model when cell-exclusive marker transcripts are erroneously assigned to a cell mask of a different type. This pushes the segmentation to respect known expression boundaries between cell types.

\paragraph{Evaluation metrics}
For evaluating the quantitative performance of the BIDCell model, we used the unmodified CellSPA framework, also developed by Fu \textit{et al.} for the BIDCell paper \cite{fu_bidcell_2024}. It captures a variety of important measures and is specifically suited to work with BIDCell. 

\section{Generative Modeling Details}
\label{S-app:generative_details}

\subsection{VAE Formulation and Protocol}
\label{S-app:vae_details}

A standard VAE \cite{Kingma2014} consists of an encoder $q_\phi(\bm{z}|\bm{x})$ that maps an input $\bm{x}$ to a latent distribution and a decoder $p_\theta(\bm{x}|\bm{z})$ that reconstructs the input from latent samples. Concretely, the encoder $\text{Enc}_\phi(\bm{x})$ produces intermediate features, which are then transformed by two heads to obtain the mean $\bm{\mu}_\phi(\bm{x})$ and log-variance $\log \bm{\sigma}_\phi^2(\bm{x})$ of a diagonal Gaussian approximate posterior,
\begin{align}
q_\phi(\bm{z}|\bm{x}) = \mathcal{N}\!\big(\bm{z};\,\bm{\mu}_\phi(\bm{x}), \mathrm{diag}(\bm{\sigma}^2_\phi(\bm{x}))\big).
\end{align}
Latent samples are drawn via the reparameterization trick: $\bm{z} = \bm{\mu}_\phi(\bm{x}) + \bm{\sigma}_\phi(\bm{x}) \odot \bm{\epsilon}$ where $\bm{\epsilon} \sim \mathcal{N}(\bm{0}, \bm{I})$. The decoder $\text{Dec}_\theta(\bm{z})$ then maps the latent sample back to the input space (e.g., as Bernoulli parameters for normalized images).

The standard evidence lower bound (ELBO) is formulated as:
\begin{equation}
    \mathcal{L}_{\text{ELBO}}(\theta,\phi) = \underbrace{\mathbb{E}_{q_\phi(\bm{z}|\bm{x})}\!\left[\log p_\theta(\bm{x}|\bm{z})\right]}_{\substack{\text{reconstruction}\\=\,-\mathcal{L}_{\text{recon}}}} - \underbrace{D_{\text{KL}}\!\left(q_\phi(\bm{z}|\bm{x}) \| p(\bm{z})\right)}_{\substack{\text{latent regularization}\\=\,\mathcal{L}_{\text{kld}}}},
\end{equation}
where $p(\bm{z})$ is typically $\mathcal{N}(\bm{0},\bm{I})$. Minimizing its negative yields two potentially conflicting terms:
(1) \emph{reconstruction quality} ($\mathcal{L}_{\text{recon}}$), implemented as a negative log-likelihood (we use BCE for image data normalized to $[0,1]$); and (2) \emph{KL divergence regularization} ($\mathcal{L}_{\text{kld}}$),
\begin{equation}
\mathcal{L}_{\text{kld}}(\bm{x}; \phi) = D_{\text{KL}}\!\left(q_\phi(\bm{z}|\bm{x}) \| p(\bm{z})\right) = \frac{1}{2}\sum_{i=1}^{d}\left[\sigma_{\phi,i}^2(\bm{x}) + \mu_{\phi,i}^2(\bm{x}) - 1 - \log\sigma_{\phi,i}^2(\bm{x})\right],
\end{equation}
which has a closed form for diagonal Gaussians. The standard (scalarized) objective in our notation is:
\begin{align}
\mathcal{L}_{\text{VAE}}(\theta, \phi) = \lambda_{\text{recon}} \mathcal{L}_{\text{recon}}(\theta, \phi) + \lambda_{\text{kld}} \mathcal{L}_{\text{kld}}(\phi),
\end{align}
where $\lambda_{\text{recon}}$ and $\lambda_{\text{kld}}$ are fixed weights (corresponding to a particular choice of $\beta$ up to scaling). These weights control the reconstruction vs./ regularization trade-offs. For example, a weak regularization can permit a mismatch between the aggregated posterior and the prior, while excessive pressure toward the prior can reduce the information carried by the latent variables in the bottleneck. 

We reformulate standard VAE training as a multi-objective optimization problem by treating reconstruction and KL divergence as separate objectives at the gradient level. Instead of computing a single gradient, we compute weighted gradients independently with $\bm{g}_{\text{recon}} = \lambda_{\text{recon}}\nabla_{\theta,\phi} \mathcal{L}_{\text{recon}}(\theta, \phi)$ and $\bm{g}_{\text{kld}} = \lambda_{\text{kld}}\nabla_{\theta,\phi} \mathcal{L}_{\text{kld}}(\phi)$, where the KL gradient has no influence in decoder's parameter. We then stack these gradients and apply gradient aggregation $\mathcal{A}$ to determine the common update direction:
\begin{align}
\bm{J}_{\text{VAE}} =
\begin{bmatrix}
\bm{g}_{\text{recon}}^\top \\
\bm{g}_{\text{kld}}^\top
\end{bmatrix},
\qquad
\bm{G}_{\text{VAE}} = \mathcal{A}(\bm{J}_{\text{VAE}}).
\end{align}

We compare LS with vanilla MGDA, its normalization variants MGDA-GN, MGDA-LN, and MGDA-LGN, the AMTL scaling modes AMTL-min and AMTL-median, and UPGrad. The MGDA and AMTL variants are defined in Appendix~\ref{S-app:agg_variants}.

\paragraph{Model architecture} We employ a convolutional VAE architecture with a latent dimension of $d=128$. The encoder consists of five convolutional layers with hidden dimensions $[32, 64, 128, 256, 512]$, followed by fully connected layers that produce the mean and log-variance of the approximate posterior. The decoder mirrors this structure with transposed convolutions to reconstruct the input image.

\paragraph{Training protocol} Models are trained for 100 epochs on CelebA and 500 epochs on CelebA-HQ, with a batch size of 128. Optimization is performed using the Adam optimizer \cite{kingma2014adam} with default momentum parameters $\beta_1=0.9$ and $\beta_2=0.999$. The learning rate is fixed at 0.001 throughout training without weight decay or learning rate scheduling.

\paragraph{Loss functions} For the CelebA results in \cref{tab:celeba_vae_results} of the main paper, reconstruction uses binary cross-entropy (BCE) with mean reduction across pixels and Bernoulli decoder parameters for images in $[0,1]$. Following the configuration in~\cite{Subramanian2020}, the weights are $\lambda_{\mathrm{recon}}=1$ and $\lambda_{\mathrm{kld}}=0.00025$. The CelebA-HQ comparison in \cref{tab:celeba_hq_vae_results_csv} of the main paper evaluates BCE and MSE with $\lambda_{\mathrm{kld}}=0.00457$; the reconstruction loss and weights are specified in each row.

\paragraph{Computational resources} All experiments are conducted on a single NVIDIA A100-SXM4-80GB GPU. Each experiment is repeated over 3 independent runs with different random seeds, and the results in \cref{tab:celeba_vae_results} of the main paper report the mean performance metrics.

\subsection{VQ-VAE and GG-VQ-VAE Formulation and Additional Results}
\label{S-app:gg_vqvae}

Vector Quantized VAEs (VQ-VAE) \cite{Oord2017} replace continuous latent variables with discrete latent codes from a learned codebook $\mathcal{E} = \{\bm{e}_k\}_{k=1}^K$. Given an encoder output $\bm{z}_e(\bm{x})$, vector quantization selects the nearest codebook entry:
\begin{align}
\bm{z}_q(\bm{x}) = \bm{e}_k, \quad \text{where } k = \arg\min_j \|\bm{z}_e(\bm{x}) - \bm{e}_j\|_2.
\end{align}
The decoder reconstructs the input from the quantized latent code, $\hat{\bm{x}} = \text{Decoder}(\bm{z}_q(\bm{x}))$.

VQ-VAE training uses three objectives: reconstruction quality ($\mathcal{L}_{\text{recon}}$), vector quantization ($\mathcal{L}_{\text{vq}}$), and commitment ($\mathcal{L}_{\text{cm}}$):
\begin{align}
\mathcal{L}_{\text{recon}}(\bm{x}, \hat{\bm{x}}; \theta, \phi) &= -\bm{x}\log\hat{\bm{x}} - (1-\bm{x})\log(1-\hat{\bm{x}}), \\
\mathcal{L}_{\text{vq}}(\bm{x}; \phi, \mathcal{E}) &= \|\text{sg}[\bm{z}_e(\bm{x})] - {\underbrace{\bm{z}_q(\bm{x})}_{\mathclap{\text{updates codebook}}}}\|_2^2, \\
\mathcal{L}_{\text{cm}}(\bm{x}; \phi, \mathcal{E}) &= \|{\underbrace{\bm{z}_e(\bm{x})}_{\mathclap{\text{updates encoder}}}} - \text{sg}[\bm{z}_q(\bm{x})]\|_2^2,
\end{align}
where $\text{sg}[\cdot]$ denotes the stop-gradient operation. The standard scalarized objective is
\begin{equation}
\mathcal{L}_{\text{VQ-VAE}}(\theta, \phi, \mathcal{E}) = \lambda_{\text{recon}} \mathcal{L}_{\text{recon}}(\theta, \phi) + \lambda_{\text{vq}} \mathcal{L}_{\text{vq}}(\phi, \mathcal{E}) + \lambda_{\text{cm}} \mathcal{L}_{\text{cm}}(\phi, \mathcal{E}).
\end{equation}
In the MOO formulation, we compute objective-wise gradients and aggregate them:
\begin{align}
G_{\text{VQ-VAE}} = \mathcal{A}([\bm{g}_{\mathrm{recon}}^\top;\bm{g}_{\mathrm{vq}}^\top;\bm{g}_{\mathrm{cm}}^\top]).
\end{align}

Here, $\bm{g}_m=\lambda_m\nabla_{(\theta,\phi,\mathcal{E})}\mathcal{L}_m$ includes the preference weight once, with zero components for parameters not affected by an objective. Gradient-Guided VQ-VAE (GG-VQ-VAE) augments the base VQ-VAE with two auxiliary objectives targeting high-frequency image structure. Let $\bm{G}(\bm{x}) = \sqrt{(\bm{S}_x * \bm{x})^2 + (\bm{S}_y * \bm{x})^2}$ denote the Sobel gradient magnitude \cite{Gonzalez2017}, and let $\bar{\bm{G}}(\bm{x})$ be its normalized version. The gradient-guided and edge-matching losses are as follows:
\begin{align}
\mathcal{L}_{\text{gg}}(\bm{x}, \hat{\bm{x}}; \theta, \phi) &= \bar{\bm{G}}(\bm{x}) \odot \mathcal{L}_{\mathrm{recon}}(\bm{x}, \hat{\bm{x}}), \\
\mathcal{L}_{\text{em}}(\bm{x}, \hat{\bm{x}}; \theta, \phi) &= \|\bm{G}(\bm{x}) - \bm{G}(\hat{\bm{x}})\|_1.
\end{align}
The five-objective GG-VQ-VAE update is then
\begin{align}
G_{\text{GG-VQ-VAE}} = \mathcal{A}([\bm{g}_{\mathrm{recon}}^\top;\bm{g}_{\mathrm{vq}}^\top;\bm{g}_{\mathrm{cm}}^\top;\bm{g}_{\mathrm{gg}}^\top;\bm{g}_{\mathrm{em}}^\top]).
\end{align}

\paragraph{Model architecture} We employ the standard VQ-VAE architecture \cite{Oord2017} with a discrete codebook of size $K=512$ and embedding dimension $d=64$. The encoder and decoder follow convolutional architectures with hidden dimensions $[128, 256]$, suitable for image generation tasks on ImageNet.

\paragraph{Training protocol} All models are trained for 20 epochs with a batch size of 128. Optimization is performed using the Adam optimizer \cite{kingma2014adam} with default momentum parameters $\beta_1=0.9$ and $\beta_2=0.999$. The learning rate is fixed at $2 \times 10^{-4}$ throughout training without weight decay or learning rate scheduling.

\paragraph{Loss functions} For the base VQ-VAE, we employ three objectives: reconstruction loss ($\mathcal{L}_{\text{recon}}$), vector quantization loss ($\mathcal{L}_{\text{vq}}$), and commitment loss ($\mathcal{L}_{\text{cm}}$). The reconstruction loss uses binary cross-entropy (BCE) with mean reduction across pixels to measure reconstruction quality. Following the original VQ-VAE formulation \cite{Oord2017}, the default weight configuration is $\lambda_{\text{recon}}=1$, $\lambda_{\text{vq}}=1$, and $\lambda_{\text{cm}}=0.25$.

The ImageNet results are reported in \cref{sec:generative} of the main paper. \Cref{S-tab:celeba_vq_vae_results,S-tab:celeba_hq_vq_vae_results} provide additional CelebA and CelebA-HQ reconstruction metrics for VQ-VAE and GG-VQ-VAE variants.

\begin{table}[ht]
    \centering
    \caption{\textbf{VQ-VAE reconstruction evaluation metrics comparison on CelebA.} We report rFID (lower is better), SSIM, SSNR, and PSNR (higher is better). The best and second-best results for each metric are shown in bold and underlined, respectively. Variants of MGDA are defined in Appendix~\ref{S-app:agg_variants}.}
    \label{S-tab:celeba_vq_vae_results}
    \resizebox{\columnwidth}{!}{%
    \begin{tabular}{@{}lccccc|cccc@{}}
    \toprule
    Method & $\lambda_{\mathrm{recon}}$ & $\lambda_{\mathrm{vq}}$ & $\lambda_{\mathrm{cm}}$ & $\lambda_{\mathrm{gg}}$ & $\lambda_{\mathrm{em}}$ & rFID $\downarrow$ & SSIM $\uparrow$ & SSNR $\uparrow$ & PSNR $\uparrow$ \\
    \midrule
    VQ-VAE + LS & 1 & 1 & 0.25 & -- & -- & 20.4939 & 0.9079 & 16.2483 & 27.7630 \\
    VQ-VAE + UPGrad & 1 & 1 & 0.25 & -- & -- & 20.8938 & 0.9093 & 16.3524 & 28.4589 \\
    \midrule
    GG-VQ-VAE + LS & 1 & 1 & 0.25 & 1 & 1 & 14.4247 & 0.9155 & 15.9929 & 27.3252 \\
    GG-VQ-VAE + LS & 1 & 1 & 0.25 & 0.25 & 1 & 12.5321 & \underline{0.9366} & 17.0780 & 28.7360 \\
    GG-VQ-VAE + LS & 1 & 1 & 0.25 & 1 & 2 & 12.2899 & 0.9304 & 16.6828 & 28.8931 \\
    \midrule
    GG-VQ-VAE + UPGrad & 1 & 1 & 0.25 & 1 & 1 & \textbf{12.1202} & \textbf{0.9376} & \textbf{17.3733} & \textbf{30.0629} \\
    GG-VQ-VAE + UPGrad & 1 & 1 & 0.25 & 0.25 & 1 & 13.1416 & 0.9299 & 16.8861 & \underline{29.8116} \\
    GG-VQ-VAE + UPGrad & 1 & 1 & 0.25 & 1 & 2 & \underline{12.1354} & 0.9362 & \underline{17.2552} & 28.8379 \\
    \bottomrule
    \end{tabular}
    }
    \end{table}

\begin{table}[ht]
    \centering
    \caption{\textbf{VQ-VAE reconstruction evaluation metrics comparison on CelebA-HQ.} We report rFID (lower is better), SSIM, LPIPS, and PSNR (higher is better for SSIM and PSNR). Within each block separated by horizontal rules, the best value per metric is \textbf{bold} and the second-best is \underline{underlined}. Variants of MGDA are defined in Appendix~\ref{S-app:agg_variants}.}
    \label{S-tab:celeba_hq_vq_vae_results}
    \resizebox{\columnwidth}{!}{%
    \begin{tabular}{@{}lccccc|cccc@{}}
    \toprule
    Method & $\lambda_{\mathrm{recon}}$ & $\lambda_{\mathrm{vq}}$ & $\lambda_{\mathrm{cm}}$ & $\lambda_{\mathrm{gg}}$ & $\lambda_{\mathrm{em}}$ & rFID $\downarrow$ & SSIM $\uparrow$ & LPIPS $\downarrow$ & PSNR $\uparrow$ \\
    \midrule
    VQ-VAE + LS & 1 & 1 & 0.25 & -- & -- & 27.1954 & 0.8547 & 0.5114 & 29.4614 \\
    VQ-VAE + UPGrad & 1 & 1 & 0.25 & -- & -- & 26.8938 & 0.9093 & 0.5012 & 30.1041 \\
    VQ-VAE + AMTL-min & 1 & 1 & 0.25 & -- & -- & \textbf{26.7128} & \textbf{0.9106} & \textbf{0.4989} & \textbf{30.2234} \\
    VQ-VAE + AMTL-median & 1 & 1 & 0.25 & -- & -- & \underline{26.8364} & \underline{0.9099} & \underline{0.5004} & \underline{30.1418} \\
    VQ-VAE + MGDA-GN & 1 & 1 & 0.25 & -- & -- & 372.3619 & 0.3836 & 1.5493 & 11.1910 \\
    VQ-VAE + MGDA-LN & 1 & 1 & 0.25 & -- & -- & 372.3619 & 0.3836 & 1.5493 & 11.1910 \\
    VQ-VAE + MGDA-LGN & 1 & 1 & 0.25 & -- & -- & 63.2621 & 0.6593 & 1.0039 & 23.0261 \\
    \midrule
    GG-VQ-VAE + LS & 1 & 1 & 0.25 & 1 & 1 & 25.4203 & \underline{0.8466} & \textbf{0.5319} & 28.5152 \\
    GG-VQ-VAE + UPGrad & 1 & 1 & 0.25 & 1 & 1 & 25.7888 & \textbf{0.8518} & \textbf{0.5319} & \textbf{28.7590} \\
    GG-VQ-VAE + AMTL-min & 1 & 1 & 0.25 & 1 & 1 & \textbf{24.2740} & 0.8428 & 0.5380 & 28.5166 \\
    GG-VQ-VAE + AMTL-median & 1 & 1 & 0.25 & 1 & 1 & \underline{25.0597} & 0.8455 & \underline{0.5357} & \underline{28.7117} \\
    GG-VQ-VAE + MGDA-LN & 1 & 1 & 0.25 & 1 & 1 & 420.2218 & 0.2738 & 1.5978 & 11.1242 \\
    GG-VQ-VAE + MGDA-LGN & 1 & 1 & 0.25 & 1 & 1 & 147.5459 & 0.5951 & 1.1784 & 21.7478 \\
    \bottomrule
    \end{tabular}
    }
    \end{table}

\section{Multi-Objective Reinforcement Learning Details}
\label{S-app:morl_details}

\subsection{Drug-Design MO-RL Protocol}
\label{S-app:drug_design_protocol}
\paragraph{Model architecture} The MO-RL model uses a Transformer encoder over SMILES fragment sequences. The base encoder has 12 layers, each using 12 self-attention heads and an embedding dimension $d=576$. After pretraining, this encoder initializes three RL components: a masker, an actor, and a critic. The critic has multiple output heads, one per objective, so that it estimates the corresponding value functions independently.

\paragraph{Pretraining protocol} The Transformer encoder is pretrained for 6 epochs on SMILES fragment sequences, alternating between masked language modeling (MLM) and contrastive learning phases, with a batch size of 100. Optimization is performed using the Adam optimizer \cite{kingma2014adam} with momentum parameters $\beta_1=0.9$ and $\beta_2=0.98$. The learning rate is fixed at 0.0001 throughout training without any learning rate scheduling.

\paragraph{Fine-tuning protocol} Given a lead molecule as the initial state, represented by a hybrid fragment-SMILES token sequence, one optimization step entails the masker selecting tokens to mask, the actor substituting the masked tokens with replacements from the token dictionary, and the critic evaluating the updated molecule sequence. This process is performed over $T=4$ timesteps, caching all molecules generated across the timesteps. At each timestep, the environment returns a separate reward signal for each objective. The multi-head critic estimates the objective-specific value functions independently, while AMTL updates the actor using the resulting objective-specific reward vectors. By contrast, the SO-RL baseline collapses the reward vector to a scalar sum before updating the policy. Once the optimization process is complete, each lead is processed sequentially to select its highest-scoring, unique optimized molecule. Fine-tuning is conducted over 150 epochs, with a batch size of 32. Model optimization is conducted using the AdamW optimizer \cite{loshchilov2017decoupled} with momentum parameters $\beta_1=0.9$ and $\beta_2=0.95$, and a numerical stability term $\epsilon=0.00001$. Fixed learning rates of $0.00003$ are used for the actor and critic, while the masker uses a learning rate of $0.00005$.

\paragraph{Loss functions} Let the masker, actor, and critic model parameters be denoted by $\theta_M$, $\theta_A$, and $\theta_C$, respectively. The advantage function, which evaluates the favourability of selecting action $a$ over other actions available at state $s$ for objective $m$ under the policy parameterized by the actor $\pi_{\theta_A}$, is defined as
\begin{equation}
\hat{A}_m^{\pi_{\theta_A}}\left(s_t, a_t\right) \approx \underbrace{r_{t,m}+\gamma r_{t+1,m}+\gamma^2 r_{t+2,m}+\cdots+\gamma^n r_{t+n,m}}_{\text{discounted rewards}} + \underbrace{\gamma^{n+1} \hat{V}_{\theta_C m}^{\pi_{\theta_A}}\left(s_{t+n+1}\right)}_{\text{bootstrapped value}}-\underbrace{\hat{V}_{\theta_C m}^{\pi_{\theta_A}}
\left(s_t\right)}_{\text{baseline}},
\end{equation}
where $r_{t,m}$ denotes the reward for objective $m$ at timestep $t$, $\gamma$ is the discount factor (set to $\gamma = 0.99$), and $\hat{V}_{\theta_C m}^{\pi_{\theta_A}}\left(s_{t}\right)$ is the state value function for objective $m$ estimated by the corresponding critic head. Letting $\pi_{\theta_M}$ denote the policy parameterized by the masker model, the loss for the masker is defined as
\begin{equation}
L\left(\theta_M\right)=\frac{1}{T} \sum_{t=1}^T \sum_{m=1}^M\left(-\hat{A}_m^{\pi_{\theta_A}}\left(s_t, a_t\right) \log \pi_{\theta_M}\left(a_t \mid s_t\right)\right).
\end{equation}
AMTL \cite{senushkin2023independent} is employed to guide the actor model, enabling balanced optimization across $M$ objectives. First, the gradient matrix $\boldsymbol{G}=\left\{\boldsymbol{g}_1, \cdots, \boldsymbol{g}_M\right\}$ is computed, where $\boldsymbol{g}_m=\nabla L_m\left(\theta_A\right)$ and 
\begin{equation}
\label{S-eq:actor_loss}
 L_m\left(\theta_A\right)=\frac{1}{T} \sum_{t=1}^T\left(-\hat{A}_m^{\pi_{\theta_{A}}}\left(s_t, a_t\right) \log \pi_{\theta_A}\left(a_t \mid s_t\right)\right).
\end{equation}    
$\boldsymbol{G}$ is then processed by the gradient matrix alignment algorithm to compute a common gradient, which is used to update the actor model. Lastly, the loss of the critic model is denoted as 
\begin{equation}
L\left(\theta_C\right)=\frac{1}{T} \sum_{t=1}^T \sum_{m=1}^M\left(r_{t,m}+\gamma\hat{V}_{\theta_C m}^{\pi_{\theta_A}}\left(s_{t+1}\right)-\hat{V}_{\theta_C m}^{\pi_{\theta_A}}\left(s_t\right)\right)^2.
\end{equation}

\paragraph{Evaluation metrics} To assess a molecule's potential as a drug candidate, a reward function is derived by considering three objectives: logP, SAS, and BAS to the LPA1 protein target, as discussed in \cref{sec:MORL} and shown in \cref{table:lead_opt} of the main paper. Additionally, validity (the ratio of optimized molecules that are chemically valid), novelty (the ratio of optimized molecules that differ from their corresponding lead molecules) and uniqueness (the ratio of unique optimized molecules) are also examined at each epoch. Results in \cref{table:lead_opt} of the main paper are based on the molecules optimized in the final epoch and averaged over 5 independent runs for MO-RL and 3 runs for all other models.

\clearpage
\bibliographystyle{IEEEtran}
\begingroup
\sloppy
\bibliography{references}

@inproceedings{parisi2014policy,
  title={Policy gradient approaches for multi-objective sequential decision making},
  author={Parisi, Simone and Pirotta, Matteo and Restelli, Marcello and Peters, Jan},
  booktitle={International Joint Conference on Neural Networks (IJCNN)},
  pages={2323--2330},
  year={2014}
}

@article{xenium_janesick_2023,
  author = {Janesick, Amanda and Shelansky, Robert and Gottscho, Andrew D. and Wagner, Florian and Williams, Stephen R. and Rouault, Morgane and Beliakoff, Ghezal and Morrison, Carolyn A. and Oliveira, Michelli F. and Sicherman, Jordan T. and Kohlway, Andrew and Abousoud, Jawad and Drber, Kashif and Shetty, Shweta and Taylor, S. Eric},
  title = {High resolution mapping of the tumor microenvironment using integrated single-cell, spatial and in situ analysis},
  journal = {Nature Communications},
  year = {2023},
  volume = {14},
  number = {1},
  pages = {8353},
  doi = {10.1038/s41467-023-43458-x}
}

@article{merfish_chen_2015,
  author = {Chen, Kok Hao and Boettiger, Alistair N. and Moffitt, Jeffrey R. and Wang, Siyuan and Zhuang, Xiaowei},
  title = {Spatially resolved, highly multiplexed {RNA} profiling in single cells},
  journal = {Science},
  year = {2015},
  volume = {348},
  number = {6233},
  pages = {aaa6090},
  doi = {10.1126/science.aaa6090}
}

@inproceedings{
  Kyriakis2022,
title={Pareto Policy Adaptation},
author={Panagiotis Kyriakis and Jyotirmoy Deshmukh and Paul Bogdan},
booktitle={International Conference on Learning Representations},
year={2022},
url={https://openreview.net/forum?id=wfZGut6e09}
}

@article{MolmaPaperYage2025,
	author = {Zhang, Yage and Zhu, Xiaodan and Li, Yifeng},
	journal = {Canadian Conference on Artificial Intelligence},
	year = {2025},
	month = {May 19},
	note = {https://caiac.pubpub.org/pub/2nguv2vy},
	publisher = {Canadian Artificial Intelligence Association (CAIAC)},
	title = {Aligning {Language} {Models} {Using} {Multi}-{Objective} {Deep} {Reinforcement} {Learning}},
}

@article{desideri2012mutiple,
title = {Multiple-gradient descent algorithm (MGDA) for multiobjective optimization},
journal = {Comptes Rendus Mathematique},
volume = {350},
number = {5},
pages = {313-318},
year = {2012},
issn = {1631-073X},
doi = {https://doi.org/10.1016/j.crma.2012.03.014},
url = {https://www.sciencedirect.com/science/article/pii/S1631073X12000738},
author = {Jean-Antoine Désidéri}
}

@article{loshchilov2017decoupled,
  title={Decoupled weight decay regularization},
  author={Loshchilov, Ilya and Hutter, Frank},
  journal={arXiv preprint arXiv:1711.05101},
  year={2017}
}

@article{yu2020gradient,
  title={Gradient surgery for multi-task learning},
  author={Yu, Tianhe and Kumar, Saurabh and Gupta, Abhishek and Levine, Sergey and Hausman, Karol and Finn, Chelsea},
  journal={Advances in Neural Information Processing Systems},
  volume={33},
  pages={5824--5836},
  year={2020}
}

@article{chen2020just,
  title={Just pick a sign: Optimizing deep multitask models with gradient sign dropout},
  author={Chen, Zhao and Ngiam, Jiquan and Huang, Yanping and Luong, Thang and Kretzschmar, Henrik and Chai, Yuning and Anguelov, Dragomir},
  journal={Advances in Neural Information Processing Systems},
  volume={33},
  pages={2039--2050},
  year={2020}
}

@article{liu2021conflict,
  title={Conflict-averse gradient descent for multi-task learning},
  author={Liu, Bo and Liu, Xingchao and Jin, Xiaojie and Stone, Peter and Liu, Qiang},
  journal={Advances in Neural Information Processing Systems},
  volume={34},
  pages={18878--18890},
  year={2021}
}

@article{navon2022multi,
  title={Multi-task learning as a bargaining game},
  author={Navon, Aviv and Shamsian, Aviv and Achituve, Idan and Maron, Haggai and Kawaguchi, Kenji and Chechik, Gal and Fetaya, Ethan},
  journal={arXiv preprint arXiv:2202.01017},
  year={2022}
}

@article{das1997closer,
  title={A closer look at drawbacks of minimizing weighted sums of objectives for Pareto set generation in multicriteria optimization problems},
  author={Das, Indraneel and Dennis, John E},
  journal={Structural optimization},
  volume={14},
  number={1},
  pages={63--69},
  year={1997},
  publisher={Springer}
}

@inproceedings{senushkin2023independent,
  title={Independent component alignment for multi-task learning},
  author={Dmitry Senushkin and Nikolay Patakin and Arseny Kuznetsov and Anton Konushin},
  booktitle={IEEE Conference on Computer Vision and Pattern Recognition (CVPR)},
  pages={20083--20093},
  year={2023}
}

@article{quinton2024jacobian,
  title={Jacobian descent for multi-objective optimization},
  author={Quinton, Pierre and Rey, Val{\'e}rian},
  journal={arXiv preprint arXiv:2406.16232},
  year={2024}
}

@inproceedings{liu2021towards,
  title={Towards impartial multi-task learning},
  author={Liu, Liyang and Li, Yi and Kuang, Zhanghui and Xue, Jing-Hao and Chen, Yimin and Yang, Wenming and Liao, Qingmin and Zhang, Wayne},
  booktitle={International Conference on Learning Representations},
  year={2021}
}

@inproceedings{lee2024parrot,
  title={Parrot: Pareto-optimal multi-reward reinforcement learning framework for text-to-image generation},
  author={Seung Hyun Lee and Yinxiao Li and Junjie Ke and Innfarn Yoo and Han Zhang and Jiahui Yu and Qifei Wang and Fei Deng and Glenn Entis and Junfeng He and Gang Li and Sangpil Kim and Irfan Essa and Feng Yang},
  booktitle={European Conference on Computer Vision},
  pages={462--478},
  year={2024},
  organization={Springer}
}

@article{Sun_2022,
  title={{Why 90\% of clinical drug development fails and how to improve it?}},
  author={Sun, Duxin and Gao, Wei and Hu, Hongxiang and Zhou, Simon},
  journal={Acta Pharmaceutica Sinica B},
  volume={12},
  number={7},
  pages={3049--3062},
  year={2022},
  publisher={Elsevier}
}

@ARTICLE{Wishart2018,
	AUTHOR = "David S. Wishart and Yannick D. Feunang and An C. Guo and Elvis J. Lo and Ana Marcu and Jason R. Grant and Tanvir Sajed and Daniel Johnson and Carin Li and Zinat Sayeeda and Nazanin Assempour and Ithayavani Iynkkaran and Yifeng Liu and Adam Maciejewski and Nicola Gale and Alex Wilson and Lucy Chin and Ryan Cummings and Diana Le and Allison Pon and Craig Knox and Michael Wilson",
	TITLE = "{DrugBank} 5.0: {A} major update to the {DrugBank} database for 2018",
	JOURNAL = "Nucleic Acids Research",
	VOLUME = {46},
	NUMBER = {Database},
	PAGES = {D1074–D1082},
	MONTH = "",
	YEAR = {2018}}

@article{TDC,
  title={Therapeutics data commons: Machine learning datasets and tasks for drug discovery and development},
  author={Huang, Kexin and Fu, Tianfan and Gao, Wenhao and Zhao, Yue and Roohani, Yusuf and Leskovec, Jure and Coley, Connor W and Xiao, Cao and Sun, Jimeng and Zitnik, Marinka},
  journal={arXiv preprint arXiv:2102.09548},
  year={2021}
}

@article{sas,
  title={{Estimation of synthetic accessibility score of drug-like molecules based on molecular complexity and fragment contributions}},
  author={Ertl, Peter and Schuffenhauer, Ansgar},
  journal={J. Cheminform.},
  volume={1},
  number={1},
  pages={1--11},
  year={2009},
  publisher={Springer}
}

@article{lin_2021,
  title={Lysophosphatidic Acid Receptor Antagonists and Cancer: {T}he Current Trends, Clinical Implications, and Trials},
  author={Lin, Yu-Hsuan and Lin, Yueh-Chien and Chen, Chien-Chin},
  journal={Cells},
  volume={10},
  number={7},
  pages={1629},
  year={2021},
  publisher={MDPI}
}

@article{schneuing_2024,
  title={Structure-based drug design with equivariant diffusion models},
  author={Arne Schneuing and Charles Harris and Yuanqi Du and Kieran Didi and Arian Jamasb and Ilia Igashov and Weitao Du and Carla Gomes and Tom Blundell and Pietro Lio and Max Welling and Michael Bronstein and Bruno Correia},
  journal={Nature Computational Science},
  volume={4},
  number={12},
  pages={899--909},
  year={2024},
  publisher={Nature Publishing Group}
}

@article{GmezBombarelli2018,
  year = {2018},
  publisher = {American Chemical Society ({ACS})},
  volume = {4},
  number = {2},
  pages = {268--276},
  author = {G{\'o}mez-Bombarelli, Rafael and Wei, Jennifer N. and Duvenaud, David and Hern{\'a}ndez-Lobato, Jos{\'e} Miguel and S{\'a}nchez-Lengeling, Benjam{\'\i}n and Sheberla, Dennis and Aguilera-Iparraguirre, Jorge and Hirzel, Timothy D. and Adams, Ryan P. and Aspuru-Guzik, Al{\'a}n},
  title = {Automatic Chemical Design Using a Data-Driven Continuous Representation of Molecules},
  journal = {{ACS} Central Science},
  DOI={10.1021/acscentsci.7b00572}
}

@article{castro2022transformer,
  title={Transformer-based protein generation with regularized latent space optimization},
  author={Castro, Egbert and Godavarthi, Abhinav and Rubinfien, Julian and Givechian, Kevin and Bhaskar, Dhananjay and Krishnaswamy, Smita},
  journal={Nature Machine Intelligence},
  volume={4},
  number={10},
  pages={840--851},
  year={2022},
  publisher={Nature Publishing Group UK London}
}

@article{Duvenaud2015,
  title={Convolutional networks on graphs for learning molecular fingerprints},
  author={Duvenaud, David K and Maclaurin, Dougal and Iparraguirre, Jorge and Bombarell, Rafael and Hirzel, Timothy and Aspuru-Guzik, Al{\'a}n and Adams, Ryan P},
  journal={Advances in Neural Information Processing Systems},
  volume={28},
  year={2015}
}

@inproceedings{Jin2018,
  title={Junction tree variational autoencoder for molecular graph generation},
  author={Jin, Wengong and Barzilay, Regina and Jaakkola, Tommi},
  booktitle={International Conference on Machine Learning},
  pages={2323--2332},
  year={2018},
  doi={10.48550/arXiv.1802.04364}
}

@article{sener_2018,
  title={Multi-task learning as multi-objective optimization},
  author={Sener, Ozan and Koltun, Vladlen},
  journal={Advances in Neural Information Processing Systems},
  volume={31},
  year={2018}
}

@article{deepFMPOv3D,
author = {Bolcato, Giovanni and Heid, Esther and Boström, Jonas},
title = {On the Value of Using {3D} Shape and Electrostatic Similarities in Deep Generative Methods},
journal = {Journal of Chemical Information and Modeling},
volume = {62},
number = {6},
pages = {1388-1398},
year = {2022}
}

@article{deepFMPO,
author = {Ståhl, Niclas and Falkman, Göran and Karlsson, Alexander and Mathiason, Gunnar and Boström, Jonas},
title = {Deep Reinforcement Learning for Multiparameter Optimization in de novo Drug Design},
journal = {Journal of Chemical Information and Modeling},
volume = {59},
number = {7},
pages = {3166-3176},
year = {2019}
}

@article{MolDQN,
  title={{Optimization of Molecules via Deep Reinforcement Learning}},
  author={Zhou, Zhenpeng and Kearnes, Steven and Li, Li and Zare, Richard N and Riley, Patrick},
  journal={Scientific Reports},
  volume={9},
  number={1},
  pages={1--10},
  year={2019},
  publisher={Nature Publishing Group}
}

@inproceedings{fu_2021,
  title={{MIMOSA}: {M}ulti-constraint Molecule Sampling for Molecule Optimization},
  author={Fu, Tianfan and Xiao, Cao and Li, Xinhao and Glass, Lucas M and Sun, Jimeng},
  booktitle={Annual AAAI Conference on Artificial Intelligence},
  volume={35},
  pages={125--133},
  year={2021}
}

@article{loeffler_2024,
  title={{Reinvent 4}: {M}odern {AI}--driven generative molecule design},
  author={Loeffler, Hannes H and He, Jiazhen and Tibo, Alessandro and Janet, Jon Paul and Voronov, Alexey and Mervin, Lewis H and Engkvist, Ola},
  journal={Journal of Cheminformatics},
  volume={16},
  number={1},
  pages={20},
  year={2024},
  publisher={Springer}
}

@InProceedings{Olaf2015,
author="Ronneberger, Olaf
and Fischer, Philipp
and Brox, Thomas",
editor="Navab, Nassir
and Hornegger, Joachim
and Wells, William M.
and Frangi, Alejandro F.",
title="{U-Net}: {C}onvolutional Networks for Biomedical Image Segmentation",
booktitle="Medical Image Computing and Computer-Assisted Intervention (MICCAI)",
year="2015",
publisher="Springer International Publishing",
address="Cham",
pages="234--241",
isbn="978-3-319-24574-4"
}

@article{fu_bidcell_2024,
  title={BIDCell: Biologically-informed self-supervised learning for segmentation of subcellular spatial transcriptomics data},
  author={Fu, Xiaohang and Lin, Yingxin and Lin, David M and Mechtersheimer, Daniel and Wang, Chuhan and Ameen, Farhan and Ghazanfar, Shila and Patrick, Ellis and Kim, Jinman and Yang, Jean YH},
  journal={Nature communications},
  volume={15},
  number={1},
  pages={509},
  year={2024},
  publisher={Nature Publishing Group UK London}
}

@article{stahl_visualization_2016,
  title={Visualization and analysis of gene expression in tissue sections by spatial transcriptomics},
  author={St{\aa}hl, Patrik L and Salm{\'e}n, Fredrik and Vickovic, Sanja and Lundmark, Anna and Navarro, Jos{\'e} Fern{\'a}ndez and Magnusson, Jens and Giacomello, Stefania and Asp, Michaela and Westholm, Jakub O and Huss, Mikael and others},
  journal={Science},
  volume={353},
  number={6294},
  pages={78--82},
  year={2016},
  publisher={American Association for the Advancement of Science}
}

@inproceedings{lin_smooth_2024,
author={Xi Lin and Xiaoyuan Zhang and Zhiyuan Yang and Fei Liu and Zhenkun Wang and Qingfu Zhang},
    title = {Smooth {T}chebycheff Scalarization for Multi-Objective Optimization},
    booktitle = {International Conference on Machine Learning},
    year = {2024}
}

@book{bishop2006pattern,
  title={Pattern recognition and machine learning},
  author={Bishop, Christopher M and Nasrabadi, Nasser M},
  volume={},
  number={},
  year={2006},
  publisher={Springer}
}

@article{Schomemann1966,
    author = {Schönemann, Peter H.},
    title = {A generalized solution of the orthogonal {P}rocrustes problem},
    journal = {Psychometrika},
    volume = {31},
    pages = {1--10},
    year = {1966}
}

@incollection{beckmann_relationship_1976,
	address = {Berlin, Heidelberg},
	title = {On the Relationship of the {T}chebycheff Norm and the Efficient Frontier of Multiple-Criteria Objectives},
	volume = {130},
	isbn = {978-3-540-07794-7 978-3-642-87563-2},
	url = {http://link.springer.com/10.1007/978-3-642-87563-2_5},
	language = {en},
	urldate = {2025-08-29},
	booktitle = {Multiple {Criteria} {Decision} {Making}},
	publisher = {Springer},
	author = {Bowman, V. Joseph},
	editor = {Beckmann, M. and Künzi, H.P. and Thiriez, H. and others},
	year = {1976},
	doi = {10.1007/978-3-642-87563-2_5},
	note = {},
	pages = {76--86},
}

@INPROCEEDINGS{Kingma2014,
     author = {D.P. Kingma and M. Welling},
      title = "Auto-encoding variational {B}ayes",
  booktitle = {International Conference on Learning Representations},
       year = {2014},
       pages= "",
   location = {},
       note = {}}

@inproceedings{Oord2017,
    author = {Aaron van den Oord and Oriol Vinyals and Koray Kavukcuoglu},
    title = {Neural Discrete Representation Learning},
    booktitle = {Conference on Neural Information Processing Systems},
    year = {2017}
}

@book{Gonzalez2017,
    author = {Rafael Gonzalez and Richard Woods},
    title = {Digital Image Processing},
    publisher = {Pearson},
    edition = {4th},
    year = {2017}
}

@inproceedings{he2016deep,
  title={Deep residual learning for image recognition},
  author={He, Kaiming and Zhang, Xiangyu and Ren, Shaoqing and Sun, Jian},
  booktitle={IEEE Conference on Computer Vision and Pattern Recognition (CVPR)},
  pages={770--778},
  year={2016}
}

@inproceedings{cui2019class,
  title={Class-balanced loss based on effective number of samples},
  author={Cui, Yin and Jia, Menglin and Lin, Tsung-Yi and Song, Yang and Belongie, Serge},
  booktitle={IEEE Conference on Computer Vision and Pattern Recognition (CVPR)},
  pages={9268--9277},
  year={2019}
}

@article{kingma2014adam,
  title={Adam: A method for stochastic optimization},
  author={Kingma, Diederik P},
  journal={arXiv preprint arXiv:1412.6980},
  year={2014}
}

@misc{Subramanian2020,
  author = {Subramanian, A.K},
  title = {PyTorch-VAE},
  year = {2020},
  publisher = {GitHub},
  journal = {GitHub repository},
  howpublished = {\url{https://github.com/AntixK/PyTorch-VAE}}
}

@article{lin2023libmtl,
  title={{LibMTL}: A {P}ython Library for Multi-Task Learning},
  author={Baijiong Lin and Yu Zhang},
  journal={Journal of Machine Learning Research},
  volume={24},
  number={209},
  pages={1--7},
  year={2023}
}

@article{zhong2024panacea,
  title={Panacea: Pareto alignment via preference adaptation for llms},
  author={Zhong, Yifan and Ma, Chengdong and Zhang, Xiaoyuan and Yang, Ziran and Chen, Haojun and Zhang, Qingfu and Qi, Siyuan and Yang, Yaodong},
  journal={Advances in Neural Information Processing Systems},
  volume={37},
  pages={75522--75558},
  year={2024}
}

@article{everson2006multiclass,
  author = {Everson, Richard M. and Fieldsend, Jonathan E.},
  title = {Multi-class {ROC} analysis from a multi-objective optimisation perspective},
  journal = {Pattern Recognition Letters},
  year = {2006},
  volume = {27},
  number = {8},
  pages = {918--927},
  doi = {10.1016/j.patrec.2005.10.016}
}
\endgroup

\end{document}